\documentclass{article}

\usepackage{PRIMEarxiv}

\usepackage[utf8]{inputenc} 
\usepackage[T1]{fontenc}    
\usepackage{url}            
\usepackage{booktabs}       
\usepackage{amsfonts}       
\usepackage{nicefrac}       
\usepackage{microtype}      
\usepackage{lipsum}
\usepackage{amsmath}
\usepackage{amsthm}
\usepackage{amssymb}   

\usepackage{algorithm}
\usepackage{algpseudocode}

\usepackage{siunitx }
\usepackage{bigints}
\usepackage{bm}
\usepackage{diagbox}
\usepackage{makecell}
\usepackage{multirow}
\usepackage{caption}
\usepackage{subcaption}
\usepackage{graphicx}

\include{pythonlisting}
\usepackage{xcolor}
\usepackage{mathtools}
\usepackage{enumitem}

\usepackage{hyperref}
\hypersetup{
    colorlinks=true,
    citecolor=black,
    linkcolor=black,
    filecolor=black,
    urlcolor=black,
}

\usepackage{pifont}
\newcommand{\cmark}{\text{\ding{51}}}
\newcommand{\xmark}{\text{\ding{55}}}

\newcommand{\R}{\mathbb{R}}

\newtheorem{theorem}{Theorem}[section]

\numberwithin{equation}{section}

\title{An Expert's Guide to Training Physics-informed Neural Networks
}

\author{
  Sifan Wang \\
Graduate Group in Applied Mathematics \\
  and Computational Science \\
  University of Pennsylvania\\
  Philadelphia, PA 19104 \\
  \texttt { sifanw@sas.upenn.edu} \\
   \And
  Shyam Sankaran \\
  Department of Mechanical Engineering \\
  and Applied Mechanics \\
  University of Pennsylvania\\
  Philadelphia, PA 19104 \\
  \texttt{shyamss@seas.upenn.edu} \\
    \AND
  Hanwen Wang \\
  Graduate Group in Applied Mathematics \\
  and Computational Science \\
  University of Pennsylvania\\
  Philadelphia, PA 19104 \\
  \texttt{wangh19@sas.upenn.edu} \\
  \And
  Paris Perdikaris \\
  Department of Mechanical Engineering \\
  and Applied Mechanics\\
  University of Pennsylvania\\
  Philadelphia, PA 19104 \\
  \texttt{pgp@seas.upenn.edu} 
}

\begin{document}
\maketitle

\begin{abstract}

Physics-informed neural networks (PINNs) have been popularized as a deep learning framework that can seamlessly synthesize observational data and partial differential equation (PDE) constraints. Their practical effectiveness however can be hampered by training pathologies, but also oftentimes by poor choices made by users who lack deep learning expertise. In this paper we present a series of best practices that can significantly improve the training efficiency and overall accuracy of PINNs. We also put forth a series of challenging benchmark problems that highlight some of the most prominent difficulties in training PINNs, and present comprehensive and fully reproducible ablation studies that demonstrate how different architecture choices and training strategies affect the test accuracy of the resulting models. We show that the methods and guiding principles put forth in this study lead to state-of-the-art results and provide strong baselines that future studies should use for comparison purposes. To this end, we also release a highly optimized library in JAX that can be used to reproduce all results reported in this paper, enable future research studies, as well as facilitate easy adaptation to new use-case scenarios.

\end{abstract}


\section{Introduction}

Recent advances in deep learning have revolutionized fields such as computer vision, natural language processing and reinforcement learning \cite{voulodimos2018deep, chowdhary2020natural, sutton2018reinforcement}. Powered by rapid growth in computational resources, deep neural networks are also  increasingly used for modeling  and simulating physical systems.
Examples of these include weather forecasting \cite{nguyen2023climax,pathak2022fourcastnet,lam2022graphcast}, quantum chemistry \cite{wang2018deepmd, pfau2020ferminet} and protein structure prediction \cite{ruff2021alphafold}. 

Notably, the fusion of scientific computing and machine learning has led to the emergence of physics-informed neural networks (PINNs) \cite{raissi2019physics}, an emerging paradigm for tackling  forward and inverse problems involving partial differential equations (PDEs). These deep learning models are known for their capability to seamlessly incorporate noisy experimental data and physical laws into the learning process. This is typically accomplished by parameterizing unknown functions of interest using deep neural networks and formulating a multi-task learning problem with the aim of matching observational data and approximating an underlying PDE system. Over the last couple of years, PINNs have led to a series of promising results across a range of problems in computational science and engineering,  including   fluids mechanics \cite{raissi2020hidden,sun2020surrogate, mathews2021uncovering}, bio-engineering    \cite{sahli2020physics, kissas2020machine}, materials \cite{fang2019deep,chen2020physics, zhang2022analyses}, molecular dynamics \cite{islam2021extraction}, electromagnetics \cite{kovacs2022conditional, fang2021high}, geosciences \cite{haghighat2021sciann, smith2022hyposvi},  and the design of thermal systems \cite{hennigh2020nvidia, cai2021physics}.

Despite some empirical success, PINNs are still facing many challenges that define open areas for research and further methodological advancements.  In recent years, there have been numerous studies focusing on improving the performance of PINNs, mostly by designing more effective neural network architectures or better training algorithms.
For example, loss re-weighting schemes have emerged as a prominent strategy for promoting a more balanced training process and  improved test accuracy \cite{wang2021understanding, wang2022and, mcclenny2020self, maddu2022inverse}. Other efforts aim to achieve similar goals by adaptively re-sampling collocation points, such as importance sampling \cite{nabian2021efficient}, evolutionary sampling \cite{daw2022rethinking} and residual-based adaptive sampling \cite{wu2023comprehensive}.
Considerable efforts have also been dedicated towards developing  new neural network architectures to improve the representation capacity of PINNs. Examples include the use of  adaptive activation functions \cite{jagtap2020adaptive}, positional embbedings \cite{liu2020multi, wang2021eigenvector}, and novel architectures \cite{wang2021understanding, sitzmann2020implicit, gao2021phygeonet, fathony2021multiplicative,moseley2021finite, kang2022pixel}.
Another research avenue explores alternative objective functions for PINNs training, beyond the weighted summation of residuals \cite{wang20222}. Some approaches incorporate numerical differentiation \cite{chiu2022can}, while others draw inspiration from Finite Element Methods (FEM), adopting variational formulations \cite{kharazmi2021hp,patel2022thermodynamically}. Other approaches propose adding  additional regularization terms to accelerate training of PINNs \cite{yu2022gradient,son2021sobolev}. Lastly, the evolution of training strategies has been an area of active research. Techniques such as sequential training \cite{wight2020solving, krishnapriyan2021characterizing} and transfer learning \cite{desai2021one, goswami2020transfer,chakraborty2021transfer}  have shown potential in speeding up the learning process and yielding better predictive accuracy.

While new research on PINNs is currently being produced at high frequency, a suite of common benchmarks and baselines is still missing from the literature. Almost all existing studies put forth their own collection of benchmark examples, and typically compare against the original PINNs formulation put forth by Raissi {\it et al.}, which is admittedly a weak baseline. This introduces many difficulties in systematically assessing progress in the field, but also in determining how to use PINNs from a practitioner's standpoint. 

To address this gap, this work  proposes a training pipeline that seamlessly integrates recent research developments to effectively resolve the identified issues in PINNs training, including spectral bias \cite{rahaman2019spectral, wang2021eigenvector}, unbalanced back-propagated gradients \cite{wang2021understanding, wang2022and} and causality violation \cite{wang2022respecting}.
In addition, we present a variety of techniques that could further enhance performance, shedding light on some practical tips that form a guideline for selecting  hyper-parameters. This is accompanied by an extensive suite of fully reproducible ablation studies performed across a wide range of benchmarks. This allows us to identify the setups that consistently yield the state-of-the-art results, which we believe should become the new baseline that future studies should compare against. We also release a high-performance library in JAX that can be used to reproduce all findings reported in this work, enable future research studies, as well as facilitate easy adaptation to new use-case scenarios. As such, we believe that this work can equally benefit researchers and practitioners to further advance PINNs and deploy them in more realistic application settings.

The rest of this paper is organized as follows. In section \ref{sec: pinns}, we provide a brief overview of the original formulation of PINNs as introduced by Raissi et al. \cite{raissi2019physics}, and outline our training pipeline. From Section \ref{sec: nondim} to Section \ref{sec: training}, we delve into  the motivation and implementation details of the  key components of the proposed algorithm. These consist of non-dimensionalization, network architectures that employ Fourier feature embeddings and random weight factorization, as well as training algorithms such as causal training, curriculum training and loss weighting strategies. Section \ref{sec: misc} discusses various aspects of PINNs that lead to improved stability and superior training performance Finally, in section \ref{sec: results} we validate the effectiveness and robustness of the proposed pipeline across a wide range of benchmarks and showcase state-of-the-art results.

\section{Physics-informed Neural Networks}
\label{sec: pinns}

Following the original formulation of Raissi {\em et al.}, we begin with a brief overview of physics-informed neural networks (PINNs) \cite{raissi2019physics} in the context of solving  partial differential equations (PDEs). Generally, we consider PDEs taking the form
\begin{align}
\label{eq: PDE}
     \mathbf{u}_t +  \mathcal{N}[\mathbf{u}] = 0, \ \  t \in [0, T],  \ \mathbf{x} \in \Omega,
\end{align} 
subject to the initial and boundary conditions
\begin{align}
     \label{eq: IC}
     &\mathbf{u}( 0, \mathbf{x})=\mathbf{g}(\mathbf{x}), \ \ \mathbf{x} \in \Omega, \\
      \label{eq: BC}
     &\mathcal{B}[\mathbf{u}] = 0,  \ \   t\in [0, T], \  \mathbf{x} \in  \partial \Omega,
\end{align}
where $\mathcal{N}[\cdot]$ is a linear or nonlinear differential operator, and  $\mathcal{B}[\cdot] $ is a boundary operator  corresponding to Dirichlet, Neumann, Robin, or periodic boundary conditions. In addition, $\mathbf{u}$ describes the unknown latent solution that is governed by the  PDE system of Equation \eqref{eq: PDE}. 

We proceed by representing the unknown solution $\mathbf{u}(t, \mathbf{x})$ by a deep neural network $\mathbf{u}_{\mathbf{\theta}}(t, \mathbf{x})$, where $\mathbf{\theta}$ denotes all tunable parameters of the network (e.g., weights and biases). This allows us to define the PDE residuals as
\begin{align}
    \label{eq: pde_residual}
    \mathcal{R}_{\mathbf{\theta}}(t, \mathbf{x}) = \frac{\partial \mathbf{u}_{\mathbf{\theta}}}{\partial t}(t_r, \mathbf{x}_r) + \mathcal{N}[\mathbf{u}_{\mathbf{\theta}}](t_r, \mathbf{x}_r)
\end{align}

Then, a physics-informed model can be trained by minimizing the following composite loss function
\begin{align}
    \label{eq: PINN_loss}
    \mathcal{L}(\mathbf{\theta}) =  \mathcal{L}_{ic}(\mathbf{\theta}) +  \mathcal{L}_{bc}(\mathbf{\theta}) +   \mathcal{L}_r(\mathbf{\theta}), 
\end{align}
where 
\begin{align}
    \label{eq: loss_ic}
     &\mathcal{L}_{ic}(\mathbf{\theta}) = \frac{1}{N_{ic}} \sum_{i=1}^{N_{ic}} \left| \mathbf{u}_{\mathbf{\theta}}(0, \mathbf{x}_{ic}^i) - \mathbf{g}(\mathbf{x}_{ic}^i) \right|^2, \\
     \label{eq: loss_bc}
     &\mathcal{L}_{bc}(\mathbf{\theta}) = \frac{1}{N_{bc}} \sum_{i=1}^{N_{bc}} \left| \mathcal{B}[\mathbf{u}_{\mathbf{\theta}}]( t_{bc}^i, \mathbf{x}_{bc}^i) \right|^2, \\
    \label{eq: loss_r}
    &\mathcal{L}_r(\mathbf{\theta}) = \frac{1}{N_r} \sum_{i=1}^{N_r} \left| \mathcal{R}_{\mathbf{\theta}}(t^i_r, \mathbf{x}^i_r) \right|^2. 
\end{align}
Here $\{\mathbf{x}_{ic}^i\}_{i=1}^{N_{ic}}$, $\{t_{bc}^i, \mathbf{x}_{bc}^i\}_{i=1}^{N_{bc}}$ and $\{t_{r}^i, \mathbf{x}_{r}^i\}_{i=1}^{N_{r}}$ can be the vertices of a fixed mesh or points that are randomly sampled at each iteration of a gradient descent algorithm. Notice that all required gradients with respect to input variables or network parameters $\mathbf{\theta}$ can be efficiently computed via automatic differentiation \cite{griewank2008evaluating}.

\begin{figure}
    \centering
    \includegraphics[width=1.0\textwidth]{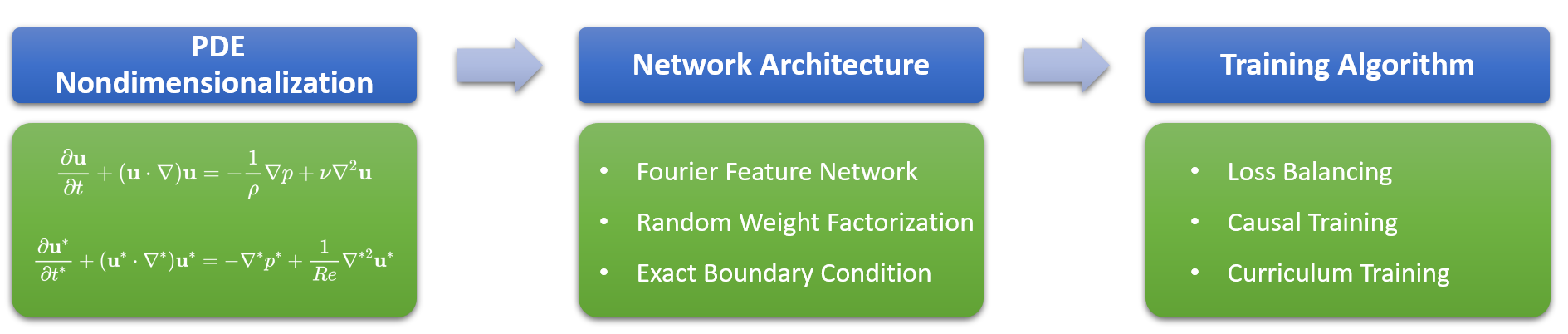}
    \caption{Illustration of the proposed training pipeline. The procedure begins with the non-dimensionalization of the PDE system, ensuring that input and output variables are in a reasonable range. Subsequently, an appropriate network architecture is constructed to represent the unknown PDE solution. The use of Fourier feature embeddings and random weight factorization is highly recommended for mitigating spectral bias and accelerating convergence. The training phase of the PINN model integrates various advanced algorithms, including self-adaptive loss balancing, causal training , and curriculum training.}
    \label{fig: pipeline}
\end{figure}

However, as demonstrated by recent work, several critical training pathologies prevent PINNs from yielding accurate and robust results. These pathologies include spectral bias \cite{rahaman2019spectral,wang2021eigenvector}, causality violation \cite{wang2022respecting}, and unbalanced back-propagated gradients among different loss terms \cite{wang2021understanding}, etc. 
To address these issues, we propose a training pipeline that integrates key recent advancements, which we believe are indispensable for the successful implementation of PINNs. As shown in Figure \ref{fig: pipeline}, the pipeline consists of three main steps, PDE non-dimensionalization, choosing suitable  network architectures and employing appropriate training algorithms. Further details are provided in Algorithm \ref{alg}. In the following sections, we will carefully demonstrate the motivation and necessity of each component in the proposed algorithm and validate its effectiveness via a wide range of benchmarks.

\begin{algorithm}
\caption{Training pipeline of physics-informed neural networks}
\label{alg: pipline}
\begin{algorithmic}
\State {1. Non-dimensionalize the PDE system \eqref{eq: PDE}.

}
\State{2. Represent the PDE solution by a multi-layer perceptron network (MLP) $\mathbf{u}_{\mathbf{\theta}}$  with Fourier feature embeddings and random weight factorization. In general, we recommend using $\tanh$ activation and initialized using the Glorot scheme.}

\State{3. Formulate the  weighted loss function according to the PDE system:
\begin{align}
    \mathcal{L}(\mathbf{\theta}) =  \lambda_{ic} \mathcal{L}_{ic}(\mathbf{\theta}) + \lambda_{bc} \mathcal{L}_{bc}(\mathbf{\theta}) +  \lambda_r \mathcal{L}_r(\mathbf{\theta}), 
\end{align}
where $\mathcal{L}_{ic}(\mathbf{\theta})$ and $\mathcal{L}_{bc}(\mathbf{\theta})$ are defined in \eqref{eq: loss_ic}, \eqref{eq: loss_bc} respectively, and 
\begin{align}
    \label{eq: splited_res_loss}
    \mathcal{L}_r(\mathbf{\theta}) = \frac{1}{M} \sum_{i=1}^M w_i \mathcal L_r^i(\mathbf{\theta}).
\end{align}
Here we partition the temporal domain into $M$ equal sequential segments and introduce $\mathcal L_r^i$ to denote the PDE residual loss within the $i$-th segment of the temporal domain. 
}

\State{4. Set all global weights $\lambda_{ic}, \lambda_{bc}, \lambda_{r}$ and temporal weights $\{w_i\}_{i=1}^M$ to 1.}

\State{5. Use $S$ steps of a gradient descent algorithm to update the parameters $\theta$ as:}

\For{$n = 1, \dots, S$} 

 \State{(a) Randomly sample $\{\mathbf{x}_{ic}^i\}_{i=1}^{N_{ic}}$, $\{t_{bc}^i, \mathbf{x}_{bc}^i\}_{i=1}^{N_{bc}}$ and $\{t_{r}^i, \mathbf{x}_{r}^i\}_{i=1}^{N_{r}}$ in the computational domain and evaluated each loss terms $ \mathcal{L}_{ic}, \mathcal{L}_{bc}$ and $\{\mathcal{L}_r^i\}_{i=1}^M$.}

  \State{(b) Compute and update the temporal weights by
  \begin{align}
  \label{eq: temporal_update}
    w_i = \exp\left(- \epsilon \sum_{k=1}^{i-1}  \mathcal{L}_r^k( \mathbf{\theta})\right)    , \text{ for } i = 2, 3, \dots, M.
  \end{align}
  Here $\epsilon > 0$ is a user-defined hyper-parameter that determines the "slope" of temporal weights.   
 }
    
 \If{$n \mod f = 0$ }

 \State{(c) Compute the global weights by
  \begin{align}
  \label{eq: lambda_ic_update}
     \hat{\lambda}_{ic} &=  \frac{  \|\nabla_{\theta}  \mathcal{L}_{ic}(\theta)\| +  \|\nabla_{\theta}  \mathcal{L}_{bc}(\theta)\| +  \|\nabla_{\theta}  \mathcal{L}_{r}(\theta)\|   } {\|\nabla_{\theta}  \mathcal{L}_{ic}(\theta)\|}, \\
     \label{eq: lambda_bc_update}
      \hat{\lambda}_{bc} &= \frac{  \|\nabla_{\theta}  \mathcal{L}_{ic}(\theta)\| +  \|\nabla_{\theta}  \mathcal{L}_{bc}(\theta)\| +  \|\nabla_{\theta}  \mathcal{L}_{r}(\theta)\|   } {\|\nabla_{\theta}  \mathcal{L}_{bc}(\theta)\|},  \\
      \label{eq: lambda_r_update}
       \hat{\lambda}_{r} &=  \frac{  \|\nabla_{\theta}  \mathcal{L}_{ic}(\theta)\| +  \|\nabla_{\theta}  \mathcal{L}_{bc}(\theta)\| +  \|\nabla_{\theta}  \mathcal{L}_{r}(\theta)\|   } {\|\nabla_{\theta}  \mathcal{L}_{r}(\theta)\|}, 
  \end{align}
where $\|\cdot\|$ denotes the $L^2$ norm.
   } 
   
\State{(d) Update the global weights $\mathbf{\lambda} = (\lambda_{ic}, \lambda_{bc}, \lambda_{r})$ using a moving average of the form
  \begin{align}
  \label{eq: lambda_update}
      \mathbf{\lambda}_{\text{new}} =  \alpha  \mathbf{\lambda}_{\text{old}}  + (1 - \alpha) \hat{\mathbf{\lambda}}_{\text{new}} .
  \end{align}}
where the parameter $\alpha$ determines the balance between the old and new values

 \EndIf
  
 \State{(e) Update the parameters $\theta$ via gradient descent
  \begin{align}
  \label{eq: theta_update}
      \theta_{n+1} &= \theta_{n} - \eta \nabla_{\mathbf{\theta}}\mathcal{L}(\mathbf{\theta}_n) \\
  \end{align}}

 \EndFor
\State{The recommended default values for hyper-parameters are as follows: $f=1,000, \alpha=0.9, \gamma=1.0, \epsilon=1.0$. Please note that we freeze the back-propagation of the weights $w_i$'s and $\lambda_i$'s with respect to network parameters $\mathbf{\theta}$.}

\end{algorithmic}
\end{algorithm}

\section{Non-dimensionalization}
\label{sec: nondim}

It is well-known that data normalization is an important pre-processing step in traditional deep learning, which typically involves scaling the input features of a data-set so that they have similar magnitudes and ranges \cite{glorot2010understanding,lecun1998gradient}. However, this process may not be generally applicable for PINNs as the target solutions are typically not available when solving forward PDE problems.  In such cases, it is important to ensure that the target output variables vary within a reasonable range. One way to achieve this is through non-dimensionalization. It is a common technique used in mathematics and physics. to simplify and analyze complex systems by transforming the original system into an equivalent dimensionless system. This is performed by selecting one or more fundamental units or characteristic values, and scaling the variables in the problem so that they become dimensionless and of order one. From our experience, non-dimensionalization plays a crucial role in building  physics-informed models especially for dealing with experimental data or real-world problems. The reasons are shown below:

\begin{itemize}
    \item Lack of consistent network initialization schemes:  The initialization of neural networks has a crucial role on the effectiveness of gradient descent algorithms. Common initialization schemes (e.g., Glorot \cite{glorot2010understanding}) not only prevent vanishing gradients but also accelerate training convergence. A critical assumption for these initialization methods is that input variables should be in a moderate range, such as having a zero mean and unit variance, which enables smooth and stable forward and backward propagation. To satisfy this assumption, we propose using non-dimensionalization to scale the input and output variables so that they are of order one. 

    \item Mitigating the disparities in variable scales: If input and output variables have different scales, some  can dominate over others, leading to unbalanced contributions in the model training, therefore hindering the learning of meaningful correlations between them. Non-dimensionalization, which scales variables to have similar magnitudes and ranges, can help to reduce this discrepancy and facilitate model training.
    
    \item Improving convergence: If the variables are not properly scaled, the optimization algorithm may have to take very small steps to adjust the weights for one variable while large steps for another variable. This may result in a slow and unstable training process. Through non-dimensionalization, the optimizer can take more consistent steps, yielding faster convergence and better performance.
\end{itemize}


While non-dimensionalization is an indispensable  pre-processing step, it is not a ``silver bullet'' that can resolve all issues in training PINNs. One of the main differences between PINNs and  conventional deep learning tasks is the minimization of PDE residuals, which introduces additional difficulties in  optimization process. Even if all variables are properly scaled via non-dimensionalization, the scale of PDE residuals can still vastly differ from the scale of the latent solution function, leading to a considerable discrepancy in the scale of different loss terms. Therefore,  it is important to carefully inspect and re-scale the loss terms that define the PINNs objective. In section \ref{sec: loss}, we introduce two self-adaptive loss weighting schemes based on the magnitude of back-propagated gradients  and Neural Tangent Kernel (NTK) theory. We will show that these methods can automatically balance the interplay between different loss terms during training and lead to more robust model performance.

\section{Network Architecture}
\label{sec: arch}

In this section, we delve into the selection of suitable network architectures for training PINNs. We begin by providing a brief  overview of multi-layer perceptrons, along with common hyper-parameter choices, activation functions, and initialization schemes. Then, we discuss random Fourier feature embeddings, a simple yet effective technique that  enables coordinate MLPs to learn  complex high frequency functions. Finally, we introduce random weight factorization, a simple drop-in replacement of dense layers that has been shown to consistently accelerate training convergence and improve model performance.

\subsection{Multi-layer Perceptrons (MLP)}

We mainly use multi-layer perceptrons (MLPs) as a universal approximator to represent the latent functions of interest,
which takes the coordinates of a spatio-temporal domain as inputs and predicts the corresponding target solution functions. Specifically,  let $\mathbf{x} \in \R^d $ be the input, $\mathbf{g}^{(0)}(\mathbf{x}) = \mathbf{x}$ and $d_0 = d$.  A MLP
$f_{\mathbf{\theta}}(\mathbf{x})$ is recursively defined by
\begin{align}
    \label{eq: mlp_1}
    \mathbf{f}^{(l)}(\mathbf{x}) = \mathbf{W}^{(l)} \cdot \mathbf{g}^{(l-1)}(\mathbf{x}) + \mathbf{b}^{(l)}, \quad \mathbf{g}^{(l)}(\mathbf{x}) = \sigma(\mathbf{f}_\theta^{(l)}(\mathbf{x})), \quad l = 1,2, \dots, L,
\end{align}
with a final layer 
\begin{align}
    \label{eq: mlp_2}
    \mathbf{f}_{\mathbf{\theta}}(\mathbf{x}) &= \mathbf{W}^{(L+1)} \cdot \mathbf{g}^{(L)}(\mathbf{x}) + \mathbf{b}^{(L+1)},
\end{align}
where $\mathbf{W}^{(l)} \in \R^{d_l \times d_{l-1}}$ is the weight matrix in $l$-th layer and $\sigma$ is an element-wise activation function. Here, $\mathbf{\theta}=\left(\mathbf{W}^{(1)}, \mathbf{b}^{(1)}, \ldots, \mathbf{W}^{(L+1)},  \mathbf{b}^{(L+1)}\right)$ 
represents all trainable parameters in the network. 

In practice, the choice of an appropriate network architecture   impacts the success of physics-informed neural networks. 
From our experience, networks that are too narrow and shallow lack the  expressive capacity to capture complex nonlinear functions, while networks that are too wide and deep can be difficult to optimize. Therefore, we recommend employing networks with width and depth ranging from $128$ to $512$ and $3$ to $6$, respectively, which tends to yield relatively optimal and robust results. To build a continuously differentiable neural representation, we recommend using the hyperbolic tangent (Tanh).   Other popular choices include  sinusoidal functions \cite{sitzmann2020implicit} and GeLU  \cite{hendrycks2016gaussian}. We point out that ReLU is not suitable since its second-order derivative is zero, which inevitably saturates the computation of PDE residuals. Finally, dense layers will be typically initialized using the Glorot scheme \cite{glorot2010understanding}.

\subsection{Random Fourier features}
\label{sec: rff}

As demonstrated by \cite{rahaman2019spectral, xu2019frequency,basri2020frequency}, MLPs suffer from a phenomenon referred to as spectral bias, showing that they are biased towards learning low frequency functions. This undesired preference also prevents PINNs from learning high frequencies and fine structures of target solutions \cite{wang2021eigenvector}. In Appendix \ref{appendix: ntk}, we present a detailed analysis of this phenomenon via a lens of Neural Tangent Kernel (NTK) theory,

To mitigate spectral bias,  Tancik {\em et al.} \cite{tancik2020fourier} proposed random Fourier feature embeddings, which map input coordinates into high frequency signals before passing through a MLP. This encoding $\mathbf{\gamma}: \mathbb{R}^n \rightarrow \mathbb{R}^m$ is defined by
\begin{align}
    \gamma(\mathbf{x})= \begin{bmatrix}
    \cos (\mathbf{B x} ) \\
    \sin (\mathbf{B x} )
    \end{bmatrix},
\end{align}
where each entry in $\mathbf{B} \in \R^{m \times d}$
is sampled from a Gaussian distribution $\mathcal{N}(0, \sigma^2)$ and $\sigma > 0$ is a user-specified hyper-parameter.

This simple method has been shown to  significantly enhance the performance of PINNs in approximating sharp gradients and complex solutions \cite{wang2021eigenvector}.
It is worth emphasizing the significance of the scale factor $\sigma$ in the performance of neural networks. As demonstrated in Appendix \ref{appendix: ntk} and \cite{wang2021eigenvector}, this hyper-parameter directly governs the frequencies of $\gamma_i$ and the resulting eigenspace of the NTK, thereby biasing the network to learn certain band-limited signals. Specifically, lower encoding frequencies can result in blurry predictions, while higher encoding frequencies can introduce salt-and-pepper artifacts. Ideally, an appropriate $\sigma$ should be selected such that the band width of NTK matches that of the target signals. This not only accelerates the training convergence, but also improves the prediction accuracy.  However, the spectral information of the solution may not be accessible when solving forward PDEs. In practice, we recommend using a moderately large $\sigma \in [1, 10]$.

\subsection{Random weight factorization}
\label{sec: rwf}

Recently, Wang {\it et al.} \cite{wang2022random} proposed random weight factorization (RWF) and demonstrated that it can consistently improve the performance of PINNs. RWF factorizes the weights associated with each neuron in the network as 
\begin{align}
    \label{eq: neuron_fact}
    \mathbf{w}^{(k, l)} =  s^{(k, l)}  \cdot \mathbf{v}^{(k, l)}, 
\end{align}
for $k=1, 2, \dots, d_l, \quad l = 1,2, \dots, L+1$, where $\mathbf{w}^{(k,l)} \in \mathbb{R}^{d_{l-1}}$ is a weight vector representing the $k$-th row of the weight matrix $\mathbf{W}^{(l)}$, $s^{(k,l)} \in \mathbb{R}$ is a trainable scale factor assigned to each individual neuron, and $\mathbf{v}^{(k,l)} \in \mathbb{R}^{d_{l-1}}$. Consequently, the proposed weight factorization can be written by
\begin{align}
    \label{eq: weight_fact}
    \mathbf{W}^{(l)} = \mathrm{diag}(\mathbf{s}^{(l)}) \cdot  \mathbf{V}^{(l)}, \quad l = 1,2, \dots, L+1.
\end{align}
with $\mathbf{s}^{(l)} \in \mathbb{R}^{d_l}$.

We provide a geometric intuition of weight factorization in Appendix \ref{appendix: rwf}. More theoretical and experimental results can be found in Appendix \ref{appendix: rwf} and \cite{wang2022random}. 

In practice, RWF is applied as follows. We first initialize the parameters of an MLP using a
standard scheme, e.g. Glorot scheme \cite{glorot2010understanding}. Then, for every weight matrix $\mathbf{W}$, we proceed by initializing a scale vector $\exp(\mathbf{s})$ where $\mathbf{s}$ is sampled from a multivariate normal distribution $\mathcal{N}(\mathbf{\mu}, \sigma \mathrm{I})$. Then every weight matrix is factorized by the associated scale factor as  $ \mathbf{W} = \mathrm{diag}(\exp(\mathbf{s})) \cdot \mathbf{V}$ at initialization.
Finally, we apply gradient descent to the new parameters $\mathbf{s}, \mathbf{V}$ directly. This procedure is summarized in Algorithm \ref{alg: rwf}. The use of exponential parameterization is motivated by Weight Normalization \cite{salimans2016weight} to strictly avoid zeros or very small values in the scale factors and allow them to span a wide range of different magnitudes. Empirically, too small $\mu, \sigma$ values may lead to performance that is similar to a conventional MLP, while too large $\mu, \sigma$ can result in an unstable training process. Therefore, we recommend setting $\mu=0.5 \text{ or } 1$,  and $\sigma=0.1$, which seem to consistently and robustly improve the loss convergence and model accuracy.

\begin{algorithm}
\caption{Random weight factorization (RWF)}\label{alg}
\label{alg: rwf}
\begin{algorithmic}
\State {1. Initialize a neural network $f_{\mathbf{\theta}}$ with $\mathbf{\theta} = \{\mathbf{W}^{(l)}, \mathbf{b}^{(l)}\}_{l=1}^{L+1}$ (e.g. using the Glorot scheme \cite{glorot2010understanding}).}
      \For{$l = 1,2, \dots, L$}
        \State {(a) Initialize each scale factor as $\mathbf{s}^{(l)}\sim\mathcal{N}(\mu, \sigma I)$.} 
        \State {(b)  Construct the factorized weight matrices as $\mathbf{W}^{(l)} = \text{diag}( \exp( \mathbf{s}^{(l)})) \cdot \mathbf{V}^{(l)}$.  }
      \EndFor
      \State {2. Train the network by gradient descent on the factorized parameters $\{\mathbf{s}^{(l)}, \mathbf{V}^{(l)}, \mathbf{b}^{(l)}\}_{l=1}^{L+1}$.}

\State {The recommended hyper-parameters are $\mu=1.0, \sigma=0.1$.} 
\end{algorithmic}
\end{algorithm}

\section{Training}
\label{sec: training}

\subsection{Respecting Temporal Causality}

\label{sec: causal}

In this section, we discuss the motivation and details of equation \eqref{eq: temporal_update} in Algorithm \ref{alg: pipline}. Recently, Wang {\em et al.} \cite{wang2022respecting} illustrates that PINNs may violate temporal causality when solving time-dependent PDEs, and hence are susceptible to converge towards erroneous solutions. This is mainly because the conventional PINNs tend to minimize all PDE residuals simultaneously meanwhile they are undesirably biased toward minimizing PDE residuals at later time, even before obtaining the correct solutions for earlier times. A more detailed analysis can be found in Appendix \ref{appendix: causality} and \cite{wang2022respecting}.

To impose the missing causal structure within the optimization process, we first split the temporal domain into $M$ equal sequential segments and introduce $L_r^i$ to denote the PDE residual loss within the $i$-th segment of the temporal domain. Then the original PDE residual loss can be rewritten as 
\begin{align}
       \mathcal{L}_r(\mathbf{\theta}) = \frac{1}{M} \sum_{i=1}^M w_i \mathcal{L}_r^i(\mathbf{\theta}).
\end{align}
Combing with equation \eqref{eq: temporal_update}, we obtain
\begin{align}
     \mathcal{L}_r(\mathbf{\theta}) = \frac{1}{M} \sum_{i=1}^M     \exp\left(- \epsilon \sum_{k=1}^{i-1}  \mathcal{L}_r^k( \mathbf{\theta})\right) \mathcal L_r^i(\mathbf{\theta}).
\end{align}
It can be observed that $w_i$ is inversely exponentially proportional to the magnitude of the cumulative residual loss from the previous time steps. As a result, $\mathcal{L}_r^i(\mathbf{\theta})$ will not be minimized unless all previous residuals $\{\mathcal{L}_r^k(\mathbf{\theta})\}_{k=1}^{i-1}$ decrease to sufficiently small value such that $w_i$ is large enough. These temporal weights encourage PINNs to the PDE solution progressively along the time axis, in accordance with how the information propagates in time, as the dynamics evolve throughout the spatio-temporal domain.

We emphasize that the computational cost of calculating temporal weights is negligible, as the temporal weights are computed by directly evaluating the PINNs loss functions, whose values are already stored in the computational graph during training. Moreover, it is important to note that the temporal weights are functions of the trainable parameters $\mathbf{\theta}$. In our JAX implementation \cite{jax2018github},  we use \texttt{lax.stop\_gradient} to avoid the computation of back-propagated gradients of temporal weights, thereby further conserving computational resources.

We must point out that the proposed weighted residual
loss does exhibit some sensitivity to the causality parameter $\epsilon$. Choosing a very small $\epsilon$ may fail to impose enough temporal causality,  resulting in the PINN model behaving similarly to the conventional one. Conversely, choosing a large $\epsilon$ value can result in a more difficult optimization problem, because the temporal residuals at earlier times have to decrease to a very small value in order to activate the latter temporal weights. Achieving this may be difficult in some cases due to limited network capacity for minimizing the target residuals. In practice, we recommend choosing a moderately large $\epsilon$ to ensure that all temporal weights can properly converge to 1 at the end of training. If this is not the case, it would be advisable to slightly reduce the value of $\epsilon$.

\subsection{Loss Balancing}

\label{sec: loss}

As mentioned in Section \ref{sec: nondim}, one of the main challenges in training PINNs is addressing multi-scale losses that arise from the minimization of PDE residuals. These losses cannot be normalized during the pre-processing step. An alternative approach involves assigning appropriate weights to each loss term to scale them during training. However, manually choosing weights is impractical, as the optimal weights can vary greatly across different problems, making it difficult to find a fixed empirical recipe that is transferable to various PDEs. More importantly, since the solution to a PDE is unknown, there is no validation data-set available for fine-tuning these hyper-parameters in the context of solving PDEs.

Given that,  our training pipeline integrates a self-adaptive learning rate annealing algorithm, which can automatically balance losses during training. 
Specifically, we first compute $\hat{\lambda}$ by equation \eqref{eq: lambda_ic_update}-\eqref{eq: lambda_r_update}. Then we obtain
\begin{align}
  \| \hat{\lambda}_{ic} \nabla_\theta \mathcal{L}_{ic} (\theta) \| =   \| \hat{\lambda}_{bc} \nabla_\theta \mathcal{L}_{ic} (\theta) \| = \| \hat{\lambda}_{r} \nabla_\theta \mathcal{L}_{ic} (\theta) \| = \| \nabla_\theta \mathcal{L}_{ic} (\theta) \| +  \| \nabla_\theta \mathcal{L}_{bc} (\theta) \|  +  \| \nabla_\theta \mathcal{L}_{r} (\theta) \| 
\end{align}
This effectively guarantees that the norm of gradients of each weighted loss is equal to each other, preventing our model from being biased towards minimizing certain terms during training. The actual weights are then updated as a running average of their previous values, as defined by Equation \eqref{eq: lambda_update}. This technique mitigates the instability of stochastic gradient descent. In practice, these updates can either take place every hundred or thousand iterations of the gradient descent loop or at a user-specified frequency. Consequently, the extra computational overhead associated with these updates is negligible, particularly when updates are infrequent.

We now introduce an alternative weighting scheme that leverages the resulting NTK matrix of PINNs \cite{wang2022and}. To this end, we define the NTK matrices corresponding to $\mathcal{L}_{ic}$, $\mathcal{L}_{bc}$, and $\mathcal{L}_{r}$ as follows:
\begin{align}
    K_{ic} &=  \left[ \left\langle \frac{\partial u}{\partial \mathbf{\theta}}(0, \mathbf{x}_{ic}^i),   \frac{\partial u}{\partial \mathbf{\theta}}(0, \mathbf{x}_{ic}^j)       \right \rangle       \right], \\
    K_{bc} &=  \left[ \left\langle \frac{\partial u}{\partial \mathbf{\theta}}(t_{bc}^i, \mathbf{x}_{bc}^i),   \frac{\partial u}{\partial \mathbf{\theta}}(t_{bc}^j, \mathbf{x}_{bc}^j)       \right \rangle       \right], \\
     K_{bc} &=  \left[ \left\langle \frac{\partial \mathcal{R}[u]}{\partial \mathbf{\theta}}(t_{r}^i, \mathbf{x}_{r}^i),   \frac{\partial \mathcal{R}[u]}{\partial \mathbf{\theta}}(t_{r}^j, \mathbf{x}_{r}^j)    \right \rangle       \right], 
\end{align}
where $\mathcal{R}[\cdot]$ denotes the residual operator defined in \eqref{eq: pde_residual}.

With this definition, we can establish an NTK-based weighting scheme as shown below
  \begin{align}
     \hat{\lambda}_{ic} &=  \frac{ Tr(\mathbf{K}_{ic}) + Tr(\mathbf{K}_{bc}) + Tr(\mathbf{K}_{r})  } {Tr(\mathbf{K}_{ic})}, \\
     \hat{\lambda}_{bc} &=  \frac{ Tr(\mathbf{K}_{ic}) + Tr(\mathbf{K}_{bc}) + Tr(\mathbf{K}_{r})  } {Tr(\mathbf{K}_{bc})}, \\
    \hat{\lambda}_{r} &=  \frac{ Tr(\mathbf{K}_{ic}) + Tr(\mathbf{K}_{bc}) + Tr(\mathbf{K}_{r})  } {Tr(\mathbf{K}_{r})}.
  \end{align}
We proceed by updating the $\lambda_i$ values using a moving average, as previously described. As detailed in Appendix \ref{appendix: ntk},  the eigenvalues of NTK characterize the convergence rate of a loss function. Higher eigenvalues imply a faster convergence rate. Given that the trace of an NTK matrix is equal to the sum of all its eigenvalues, this scheme aims to balance the convergence rates of different loss terms such that their convergence rates are comparable to one another. In practice, it should be noted that we avoid constructing the full NTK matrix. Instead, we evaluate only the diagonal elements of the NTK matrix for computing the weights, which significantly saves computational resources. 

We observed that while the performance of the gradient-based and NTK-based weighting schemes is similar, the updated weights in the gradient-based scheme are less stable compared to the NTK-based scheme. This instability may be attributed to the noisy back-propagated gradients due to  random mini-batches. However, the NTK-based scheme demands a higher computational cost, making it more difficult to scale to complex problems. As a result, we generally recommend employing the gradient-based scheme as a first choice.

\subsection{Curriculum Training}


While the techniques detailed in the preceding sections have greatly enhanced the performance and application range of PINNs, there remain certain complex domains where PINNs encounter challenges, especially in scenarios where high predictive accuracy is required. For example, when simulating chaotic dynamical systems such as the Navier-Stokes equations at high Reynolds numbers, enhanced accuracy is required to prevent error accumulation and trajectory divergence. In this section, we aim to shed light on these challenging areas and explore pathways to overcome them.

A promising approach we will delve into is the curriculum training strategy introduced by Krishnapriyan {\em et. al.} \cite{krishnapriyan2021characterizing}. The core idea involves decomposing the entire optimization task for PINNs into a sequence of more manageable sub-tasks. In this work, we mainly focus on  integrating this strategy into our training pipeline for solving time-dependent PDEs and singular perturbation problems.

For time-dependent PDEs, we divide  the temporal domain into smaller intervals and employ Algorithm \ref{alg} to train PINNs for solving the PDE within each of these intervals.  Except for the first time window, the initial condition for each subsequent time window is updated using the prediction from the last time-step of the previous time window.  This approach can be viewed as a temporal domain decomposition strategy, and significantly reduces the optimization difficulty of learning the full evolution of a dynamical system while increasing computational costs due to model retraining for each window.

It is worth noting that we also partition the temporal domain in Algorithm \ref{alg} to compute the causal weights within the time-window. We emphasize that the causal weighting shares a similar motivation with ``time-marching'', in the sense of respecting temporal causality by learning the solution sequentially along the time axis. Nevertheless, the causal weighting discussed in section \ref{sec: causal} should not be considered a replacement for time-marching approaches, but rather a crucial enhancement, as violations of causality may still occur within each time window of a time-marching algorithm.


In addressing singular perturbation problems, our strategy hinges on a progressive approach. We initiate the training process
with a less singular PDE and progressively increase its singularity throughout the training. For example, if our goal is to solve the Navier-Stokes equation at moderately high Reynolds numbers, we typically start by training a model for a lower Reynolds number and use this result as a suitable initialization for minimizing PDE residuals at higher Reynolds numbers. Through our experiments, we have observed that this approach effectively stabilizes the training process. It reduces the likelihood of PINNs  becoming trapped in unfavorable local minima, thus enabling them to accurately capture complex and nonlinear PDE solutions. For a more concrete illustration, readers are directed to the example of lid-driven cavity flow in section \ref{sec: ldc}.

\section{Miscellaneous}
\label{sec: misc}

In this section, we introduce several aspects that researchers and practitioners should consider when using PINNs to promote robust and optimal performance. The discussion highlights the importance of selecting appropriate optimizers and learning rates, imposing exact boundary conditions, employing random sampling and a modified MLP architecture. 

\subsection{Optimizer and learning rate}

Numerous optimizers have been developed for deep learning applications; however, we find that the Adam optimizer consistently yields good performance without heavy tuning. Moreover, we do not recommend using weight decay especially for  forward problems, as it  tends to degrade the resulting predictive accuracy. Furthermore, the learning rate is a crucial factor in PINNs' performance. Our experience suggests that an initial learning rate of $0.001$, combined with exponential decay, typically yields good results.

\subsection{Random sampling}

The choice of an appropriate sampling strategy for collocation points plays an important role in the training efficiency and model performance. In comparison to full batch sampling,  random sampling significantly reduces the  memory requirements and the computational cost of each iteration. More importantly, random sampling introduces regularization effects, which ultimately contribute to the improved generalization capabilities of PINNs. Based on our observations, training PINNs using full-batch gradient descent may result in over-fitting the PDE residuals. Consequently, we strongly recommend using random sampling in all PINN simulations to achieve optimal performance.

\subsection{Imposing boundary conditions}

\label{sec: periodic_bc}

Recent work by Dong  {\it et al.} \cite{dong2021method} showed how to strictly impose periodic boundary conditions in PINNs as hard-constraints, which not only effectively reduces the number of loss constraints but also significantly enhances training convergence and predictive accuracy. To illustrate the main idea, let us consider enforcing periodic boundary conditions with period $P$ in a one-dimensional setting. Our goal is to build a network architecture satisfying 
\begin{align}\label{eq:periodic_constraint}
    u^{(l)}(a) = u^{(l)}(a + P), \quad l=0, 1, 2, \dots.
\end{align}
To this end, we construct a special Fourier feature embedding of the form
\begin{align}
    \label{eq: 1D_Fourier}
    \mathbf{v}(x) = \left(\cos (\omega x), \sin (\omega x) \right),
\end{align}
with $\omega = \frac{2 \pi}{L}$. Then, for any network representation $u_{\mathbf{\theta}}$, it can be proved that any $u_{\mathbf{\theta}}(v(x))$ exactly satisfies the periodic boundary condition. 

The same idea can be directly extended to higher-dimensional domains. For instance, let $(x, y)$ denote the coordinates of a point in two dimensions, and suppose that $u(x, y)$ is a smooth periodic function to be approximated in a periodic cell $[a, a + P_x] \times [b, b + P_y]$, satisfying the following constraints 
\begin{align}
    &\frac{\partial^{l}}{\partial x^{l}} u\left(a, y\right)=\frac{\partial^{l}}{\partial x^{l}} u\left(a + P_x, y\right), \quad  y \in\left[b, b + P_y\right], \\
    &\frac{\partial^{l}}{\partial y^{l}} u\left(x, a\right)=\frac{\partial^{l}}{\partial y^{l}} u\left(x, b + P_y\right), \quad  x \in\left[a, a + P_x\right],
\end{align}
for $l=0, 1, 2, \dots$, where  $P_x$  and $P_y$ are the periods in the $x$ and $y$ directions, respectively. Similar to the one-dimensional setting, these constraints can be implicitly encoded in a neural network by constructing a two-dimensional Fourier features embedding as
\begin{align}
    \mathbf{v}(x, y) = \begin{bmatrix}
    \cos \left(\omega_{x} x\right), \sin \left(\omega_{x} x\right), \cos \left(\omega_{y} y\right),  \sin \left(\omega_{y} y\right)
    \end{bmatrix}
\end{align}
with $w_x = \frac{2 \pi}{P_x}, w_y = \frac{2 \pi}{P_y}$.

For time-dependent problems, we simply concatenate the time coordinates $t$ with the constructed Fourier features embedding, i.e., $u_{\mathbf{\theta}}([t, \mathbf{v}(x)])$,  or $u_{\mathbf{\theta}}([t, \mathbf{v}(x, y)])$.  In particular, if the PDE solutions are known to exhibit periodic behavior over time, we can also enforce periodicity along the time axis. More precisely, we consider the following special Fourier embedding
\begin{align}
    \mathbf{w}(t, x) = [\cos(\omega_t t), \sin(\omega_t t), \mathbf{v}(t, x)]
\end{align}
where $\omega_t = \frac{2 \pi}{P_t}$. The key difference is that $P_t$ is a trainable parameter. 
Typically,  $P_t$ is initialized to the temporal domain's length, allowing networks to learn the solution's correct period. It is worth emphasizing that this assumption of time periodicity is not a severe restriction, and this technique can be applied to arbitrary dynamical systems, even if the solution is not periodic. This is because one can always set the initial $P_t$ greater than the length of the temporal domain.

Lastly, other types of boundary conditions, including  Dirichlet, Neumann, Robin, etc., can also be enforced in a ``hard'' manner by modifying the network outputs, see \cite{sukumar2021exact, lu2021physics} for more details.

\subsection{Modified MLP}

In practice, we found that a simple modification of MLPs proposed by Wang {\em et al.} \cite{wang2021understanding}  demonstrates an enhanced capability for learning nonlinear and complex PDE solutions. The forward pass of an $L$-layer modified MLP is defined as follows. First, we introduce two encoders for the input coordinates
\begin{align}
    &\mathbf{U} = \sigma(   \mathbf{W}_1 \mathbf{x} + \mathbf{b}_1), \ \  \mathbf{V} = \sigma( \mathbf{W}_2 \mathbf{x}  + \mathbf{b}_2).
\end{align}
Then, for $ l = 1,2, \dots, L$, 
\begin{align}
   \mathbf{f}^{(l)}(\mathbf{x}) &= \mathbf{W}^{(l)} \cdot \mathbf{g}^{(l-1)}(\mathbf{x}) + \mathbf{b}^{(l)}, \\ 
    \mathbf{g}^{(l)}(\mathbf{x}) &= \sigma(\mathbf{f}_\theta^{(l)}(\mathbf{x})) \odot \mathbf{U} + (1 - \sigma(\mathbf{f}_\theta^{(l)}(\mathbf{x}))) \odot \mathbf{V}.
\end{align}
The final network output is given by
\begin{align}
        \mathbf{f}_{\mathbf{\theta}}(\mathbf{x}) &= \mathbf{W}^{(L+1)} \cdot \mathbf{g}^{(L)}(\mathbf{x}) + \mathbf{b}^{(L+1)}.
\end{align}
Here, $\sigma$ denotes a nonlinear activation function, and $\odot$ denotes an element-wise multiplication.  All trainable parameters are given by
\begin{align}
    \mathbf{\theta} = \{\mathbf{W}_1, \mathbf{b}_1, \mathbf{W}_2, \mathbf{b}_2, (\mathbf{W}^{(l)}, \mathbf{b}^{(l)})_{l=1}^{L+1}\}.
\end{align}
This architecture is almost the same as a standard MLP network, with the addition of two encoders and a minor modification in the forward pass.  Specifically, the inputs $\mathbf{x}$ are embedded into a feature space via two encoders $\mathbf{U}, \mathbf{V}$, respectively, and merged in each hidden layer of a standard MLP  using a point-wise multiplication. In our experience, the modified MLP demands greater computational resources; however, it generally outperforms the standard MLP in effectively minimizing PDE residuals, thereby yielding more accurate results.

\section{Results}
\label{sec: results}

In this section, we present a series of challenging benchmarks for evaluating PINNs performance and illustrate the effectiveness of Algorithm \ref{alg: pipline}, along with the proposed training strategies. Besides, we showcase the state-of-the-art results for each benchmark, demonstrating the current performance capacity of PINNs. More importantly, we believe that these results can establish robust and strong baselines, enabling future researchers to perform thorough evaluations and comparisons of their novel methods. This paves the way for continued innovation and developments in this field.

For each benchmark, except for the last two,  we perform comprehensive ablation studies to assess the effectiveness of the methods presented in the previous sections. In each ablation study we systematically disable each methodological component individually, while keeping the others active under the same hyper-parameter settings, and evaluate the resulting relative $L^2$ error and run-time. This allows us to isolate the effects of each component and understand their contribution to the overall model performance.  Throughout all ablation studies, we  maintain the following hyper-parameter settings, unless stated otherwise. Specifically, we employ an MLP with 4 hidden layers, 256 neurons per hidden layer, and tanh activation functions as our backbone,  initializing it using the Glorot scheme \cite{glorot2010understanding}. 
For model training, we use the Adam optimizer \cite{kingma2014adam}, starting with a learning rate of $10^{-3}$ and  an exponential decay with a decay rate of 0.9 for every $2,000$ decay steps. The collocation points are uniformly sampled from the computational domain with a batch size of $4096$.  The total number of training iterations can vary depending on the complexity of the example.

Furthermore, we conduct extensive hyper-parameter sweeps across various learning rate schedules, network sizes, and activations, in order to produce state-of-the-art results for each example. Note that the hyper-parameter settings for our ablation studies differ from those yielding the best results. We summarize our results in Table \ref{tab: sota} and provide detailed hyper-parameter settings for our optimal models in the Appendix. Throughout all numerical experiments, when applicable, we enforce the exact periodic boundary conditions as described in section \ref{sec: misc}. 

\begin{figure}
    \centering
    \includegraphics[width=0.4\textwidth]{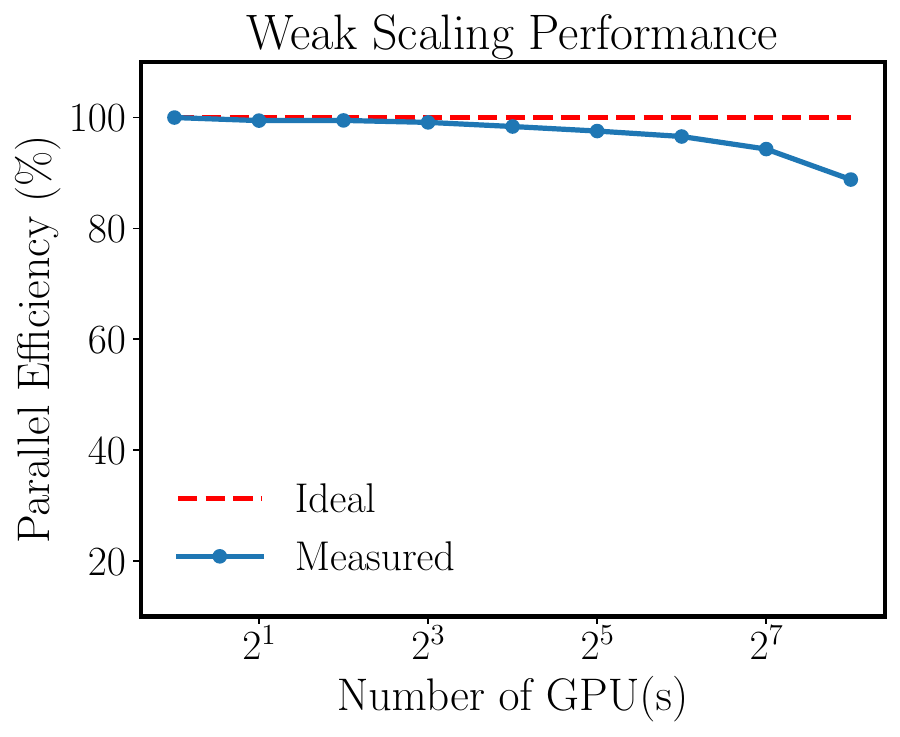}
    \caption{Efficiency of weak scaling using the Navier-Stokes flow (section \ref{sec: ns_tori}) as a benchmark. We employ a neural network with hyper-parameters shown in Table \ref{tab: ns_tori_config} and measure the execution time for 10,000 iterations, maintaining a consistent batch size of 40960 per GPU.}
    \label{fig: weak_scaling}
\end{figure}

The code and data accompanying this manuscript will be made publicly available at \url{https://github.com/PredictiveIntelligenceLab/jaxpi}. It should be highlighted that our implementation automatically supports efficient data-parallel multi-GPU training. As illustrated in Figure \ref{fig: weak_scaling}, we show great weak scaling capabilities up to 256 GPUs,  enabling the effective simulation of large-scale problems. Additionally, our code includes valuable utilities for monitoring gradient norms and NTK eigenvalues throughout training—key metrics essential for identifying potential training issues.

\begin{table}
    \centering
    \begin{tabular}{lc}
        \toprule
        \textbf{Benchmark} & \multicolumn{1}{c}{\textbf{Relative $L^2$ error}} \\
        \midrule
        Allen-Cahn equation & \num{5.37e-5} \\
        Advection equation & \num{6.88e-4} \\
        Stokes flow & \num{8.04e-5} \\
        Kuramoto–Sivashinsky equation & \num{1.61e-1} \\
        Lid-driven cavity flow (Re=3200) & \num{1.58e-1} \\
        Navier–Stokes flow in a torus & $2.45 \times 10^{-1}$ \\
        Navier–Stokes flow around a cylinder & -- \\
        \bottomrule
    \end{tabular}
    \caption{State-of-the-art relative $L^2$ error for various benchmark equations using our proposed model.}
    \label{tab: sota}
\end{table}

\subsection{Allen-Cahn equation}

We start with 1D Allen-Cahn equation, a representative case with which conventional PINN models are known to struggle. It takes the form 
\begin{align}
    &u_{t}-0.0001 u_{x x}+5 u^{3}-5 u=0, \quad t \in[0,1], x \in[-1,1], \\
    &u(0, x)=x^{2} \cos (\pi x), \\
    &u(t, -1)=u(t, 1), \\
    &u_{x}(t, -1)=u_{x}(t, 1).
\end{align}

\begin{figure}
    \centering
 \includegraphics[width=0.7\textwidth]{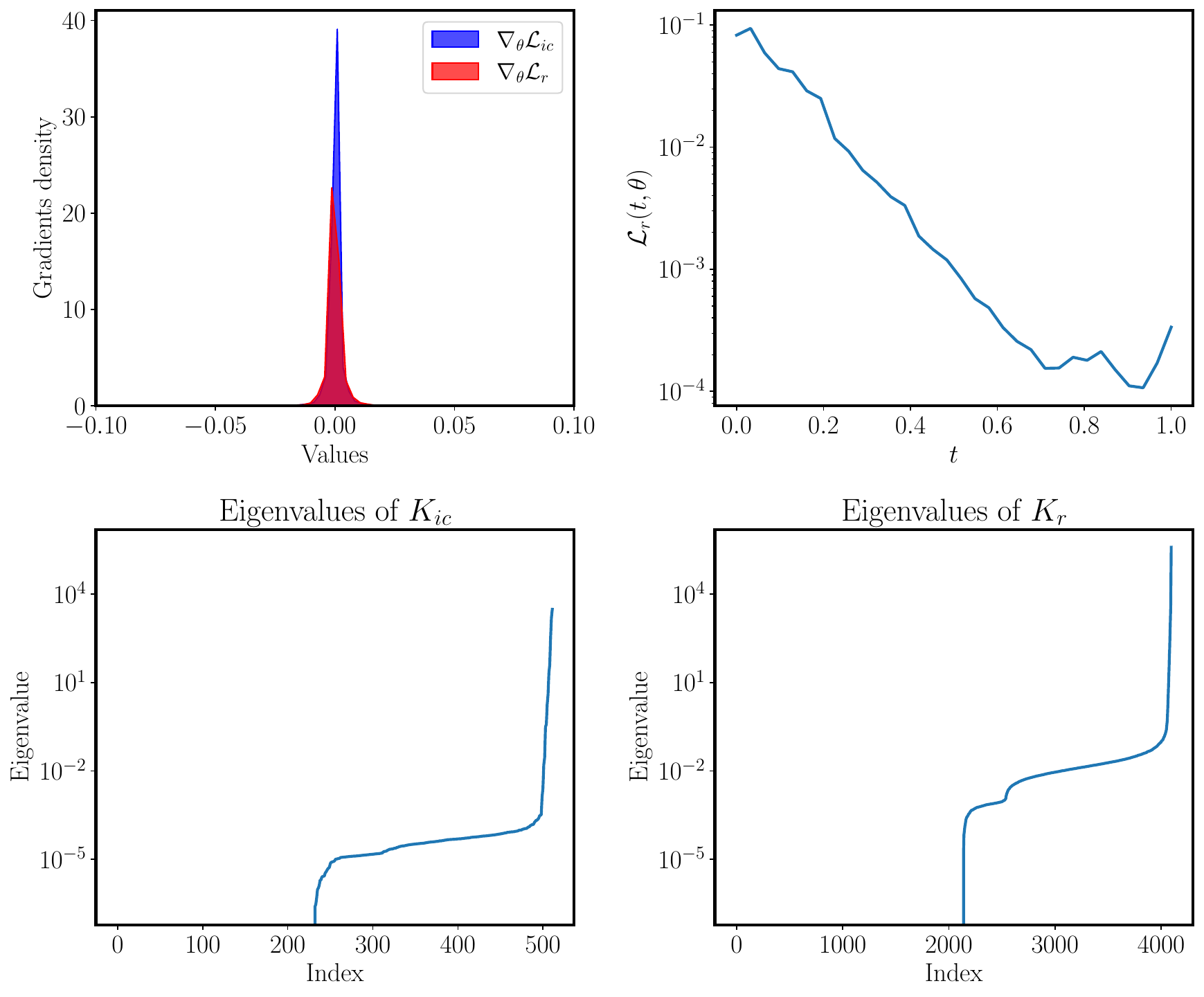}
    \caption{{\em Allen Cahn equation:} Analysis of training a plain PINN model for $10,000$ iterations.
    {\em Top left}: Histograms of back-propagated gradients of the PDE residual loss and initial condition loss at the last iteration. {\em Top right}: Temporal PDE residual loss at the last iteration. {\em Bottom:} NTK eigenvalues of $K_{ic}$ and $K_r$ at the last iteration.}
    \label{fig: ac_stat}
\end{figure}

For this example, we first train a conventional PINN model to diagnose potential issues. In Figure \ref{fig: ac_stat}, we  visualize the histogram of back-propagated gradients, the resulting NTK eigenvalues  and the temporal PDE residual loss (equation \eqref{eq: splited_res_loss}) at the early stages of training. On the top left panel, one can see that the gradients of PDE residual loss dominates those of  the initial condition loss, which implies unbalanced back-propagated gradients. Moreover, the top right panel reveals that the network tends to minimize the  PDE residuals at later times first, suggesting a violation of causality.  In the bottom panel, a rapid decay in the NTK eigenvalues can be observed, indicating the presence of spectral bias. These findings strongly suggest that conventional PINNs suffer from multiple severe training pathologies,  which need to be addressed simultaneously to yield satisfactory results.

\begin{figure}
    \centering
 \includegraphics[width=0.5\textwidth]{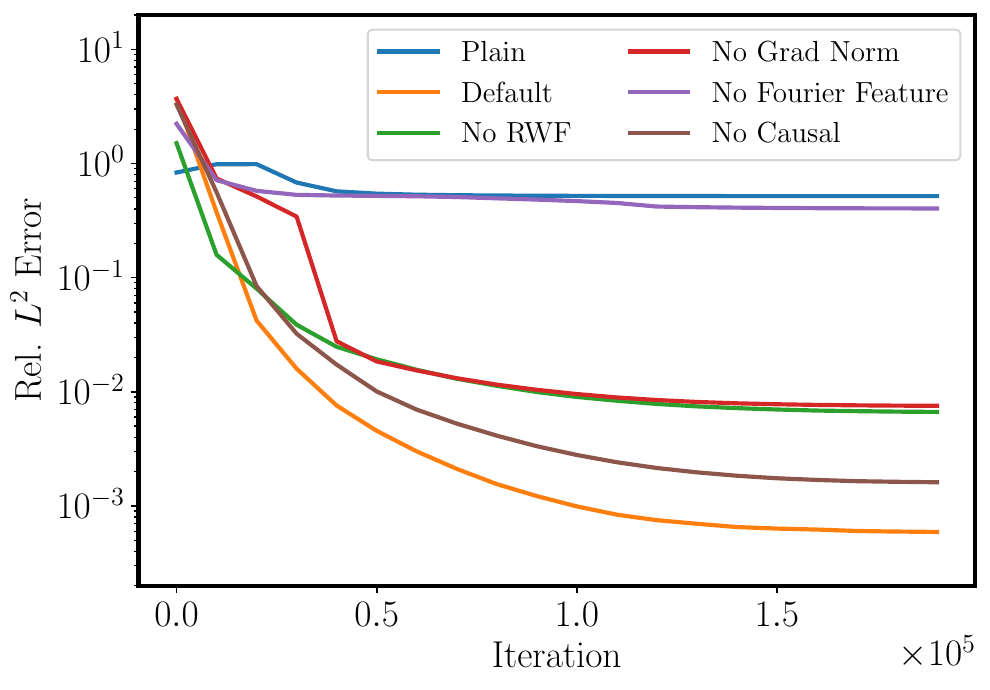}
    \caption{{\em Allen Cahn equation:}  Convergence of relative $L^2$ error for the ablation study with different components disabled. Plain: Conventional PINN formulation. Default: PINN model trained using Algorithm \ref{alg: pipline}. No RWF: PINN model trained using Algorithm \ref{alg: pipline} without random weight factorization. No Grad Norm: PINN model trained using Algorithm \ref{alg: pipline} without grad norm weighting scheme. No Fourier feature: PINN model trained using Algorithm \ref{alg: pipline} without random Fourier feature embeddings. No Causal: PINN model trained using Algorithm \ref{alg: pipline} without casual weighting.}
    \label{fig: ac_ablation}
\end{figure}

\begin{figure}
    \centering
    \includegraphics[width=0.9\textwidth]{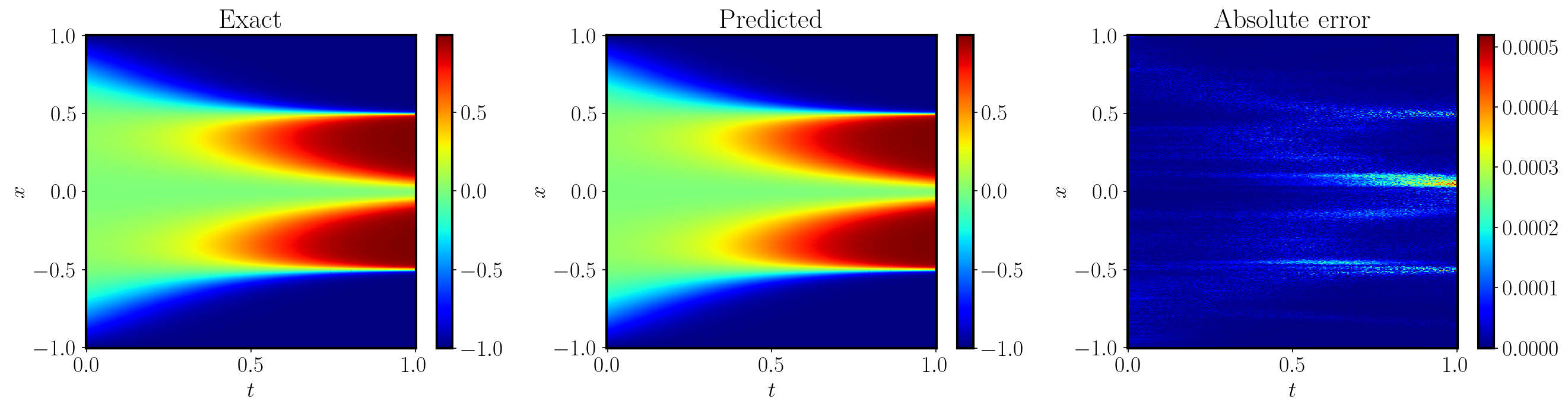}
    \caption{{\em Allen Cahn equation:} Comparison of the best prediction against the reference solution. The resulting relative $L^2$ error is $5.37 \times 10^{-5}$.  The hyper-parameter configuration can be found in Table \ref{tab: ac_config}.}
    \label{fig: ac_pred}
\end{figure}

To showcase the effectiveness of the proposed training pipeline in addressing these issues, we employ Algorithm \ref{alg: pipline} and disable individual methodological components one-at-a-time.
The results are summarized in Table \ref{tab: ac} and Figure \ref{fig: ac_ablation}. It can be concluded that the full algorithm yields the best performance in terms of relative $L^2$ error of $5.84 \times 10^{-4}$. Removing any individual component from the algorithm generally leads to a worse performance, which indicates the positive contributions of each component to the overall model performance. The most significant negative impact on performance occurs when disabling the Fourier Feature embedding, resulting in a relative $L^2$ error of $4.35 \times 10^{-1}$. It implies that the spectral bias degrades the predictive accuracy the most for this example. Furthermore, it is worth noting that the run-times across different configurations are relatively similar, except for the case corresponding to conventional PINNs, which shows a slightly shorter run-time of $12.93$ minutes. This highly suggests the computational efficiency of each component presented in Algorithm \ref{alg: pipline}. 
Finally, we present our best result in Figure \ref{fig: ac_pred}, whereas Table \ref{tab: ac_config} details the corresponding hyper-parameter configuration, and Figure \ref{fig: ac_loss_weights} visualizes the loss convergence and the weight changes during training.
One can see that the predicted solution achieves an excellent agreement with the reference solution,  yielding a relative $L^2$ error of $5.37 \times 10^{-5}$.

\begin{table}[]
    \centering
    \renewcommand{\arraystretch}{1.2}
    \begin{tabular}{ccccccc}
    \toprule
    \multicolumn{4}{c}{\textbf{Ablation Settings}} & \multicolumn{2}{c}{\textbf{Performance}} \\ 
    \cmidrule(lr){1-4} \cmidrule(lr){5-6}
    \textbf{Fourier Feature} & \textbf{RWF} & \textbf{Grad Norm} & \textbf{Causal} & \textbf{Rel. $L^2$ error} & \textbf{Run time (min) } \\ 
    \midrule
    $\cmark$ & $\cmark$ & $\cmark$ & $\cmark$ & $5.84 \times 10^{-4}$ & 16.26 \\ 
    $\xmark$ & $\cmark$ & $\cmark$ & $\cmark$ &  $4.35 \times 10^{-1}$ & 13.20 \\ 
    $\cmark$ & $\xmark$ & $\cmark$ & $\cmark$ &  $6.62 \times 10^{-3}$ & 16.53 \\ 
    $\cmark$ & $\cmark$ & $\xmark$ & $\cmark$ &  $7.51 \times 10^{-3}$ & 16.36\\ 
    $\cmark$ & $\cmark$ & $\cmark$ & $\xmark$ &  $1.59 \times 10^{-3}$ & 16.11 \\ 
    $\xmark$ & $\xmark$ & $\xmark$ & $\xmark$ &  $51.74 \times 10^{-1}$ & 12.93 \\ 
    \bottomrule
    \end{tabular}
    \caption{{\em Allen Cahn equation:} Relative $L^2$ error and run time for an ablation study illustrating the impact of disabling individual components of the proposed training pipeline. 
    Note that the GPU run time may vary due to factors such as hardware utilization, batch processing, and other computational loads.}
    \label{tab: ac}
\end{table}

\subsection{Advection equation}

Our second example is a 1D advection equation;  a linear hyperbolic equation commonly used to model transport phenomena. It takes the form
\begin{align}
    \frac{\partial u}{\partial t}+ c \frac{\partial u}{\partial x} &=0, \quad  t \in[0, 1],  \ x \in (0, 2 \pi),\\
    u(0, x) &=g(x), \quad x \in (0, 2 \pi),
\end{align}
with periodic boundary conditions. This example has been studied in \cite{krishnapriyan2021characterizing, daw2022rethinking}, exposing some of the limitations that PINNs suffer from as the transport velocity $c$ is increased. In our experiments, we consider the challenging setting of $c=80$ with an initial condition $g(x) = \sin(x)$.

Analogous to the first example, we train a conventional PINN model with the aim of identifying the issues that lead to inaccurate results. As illustrated in Figure \ref{fig: ac_stat}, it is evident that  PINNs experience the same challenges as those observed in the first example.   This observation strongly suggests the widespread nature of these issues in the training of PINNs, further emphasizing the necessity of addressing them to obtain robust and accurate PINN models.

\begin{figure}
    \centering
    \includegraphics[width=0.9\textwidth]{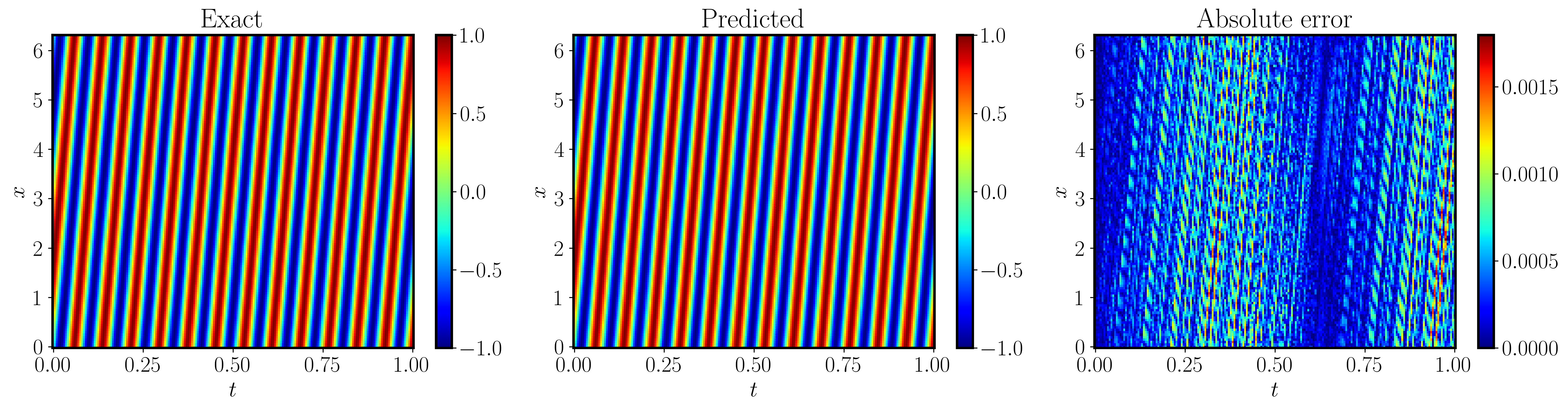}
    \caption{{\em Advection equation:}  Comparison of the best prediction against the reference solution obtained from the hyper-parameter sweep. The resulting relative $L^2$ error is $6.88 \times 10^{-4}$. The hyper-parameter configuration can be found in Table \ref{tab: adv_config}.}
    \label{fig: adv_pred}
\end{figure}

As mentioned in section \ref{sec: periodic_bc}, we can impose the spatial and temporal periodicity by 
\begin{align}
    \mathbf{v}(t, x) = [\cos(\omega_t t), \sin(\omega_t t), \cos(\omega_x x), \sin(\omega_x x)],
\end{align}
where  $\omega_t = \frac{2 \pi}{P_t}$ and $\omega_x = \frac{2 \pi}{P_x}$ with $P_x = 2 \pi$ and $P_t$ a trainable parameter.  For this example, we incorporate the imposition of temporal periodicity in Algorithm \ref{alg: pipline} and subsequently  perform an ablation study on the enhanced algorithm. The performance of various configurations is summarized in Table \ref{tab: adv}. One can conclude that the integration of all the techniques together yields the optimal accuracy. The exclusion of any of these elements, especially the time periodicity, Fourier Features and the grad norm weighting scheme, leads to a significant increase in test error, highlighting their crucial role in achieving accurate results. Additionally, we present the state-of the-art result in Figure \ref{fig: adv_pred}. We see that the model prediction achieves an excellent agreement with the exact solution, with an relative $L^2$ error  of $6.88 \cdot 10^{-4}$.  The hyper-parameter configuration and loss convergence are presented in Table \ref{tab: adv_config} and Figure \ref{fig: adv_loss_weights}, respectively.

\begin{figure}
    \centering
    \includegraphics[width=0.8\textwidth]{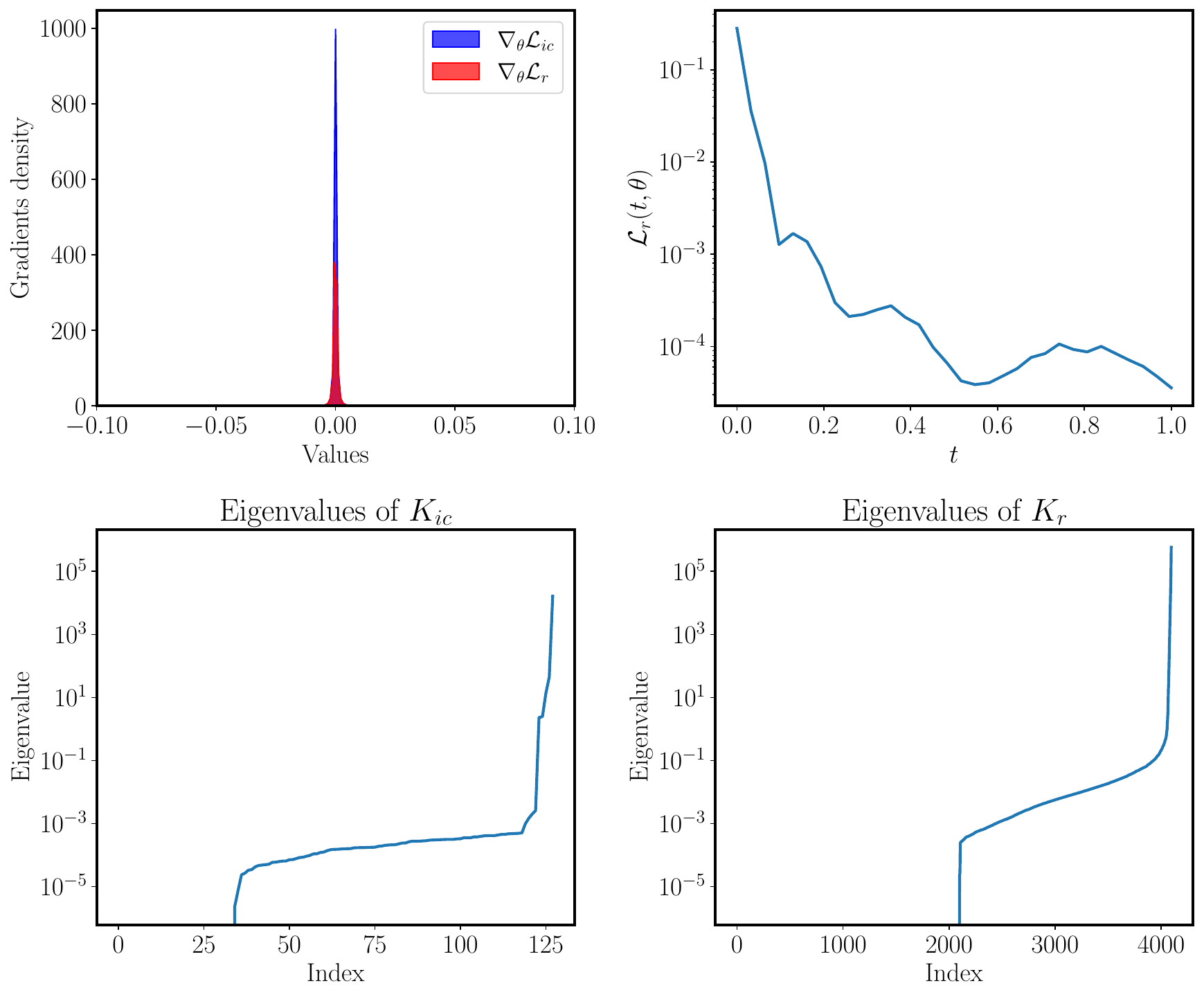}
    \caption{{\em Advection equation:}  Analysis of training a plain physics-informed neural network for 10,000 iterations. {\em Top left}: Histograms of back-propagated gradients of the PDE residual loss and initial condition loss at the last iteration. {\em Top right}: Temporal PDE residual loss  at the last iteration. {\em Bottom:} NTK eigenvalues of $K_{ic}$ and $K_r$ at the last iteration.}
    \label{fig: adv_stat}
\end{figure}

\begin{table}[]
    \centering
    \renewcommand{\arraystretch}{1.2}
    \begin{tabular}{ccccccc}
    \toprule
    \multicolumn{5}{c}{\textbf{Ablation Settings}} & \multicolumn{2}{c}{\textbf{Performance}} \\ 
    \cmidrule(lr){1-5} \cmidrule(lr){6-7}
    \textbf{Time Period} & \textbf{Fourier Feature} & \textbf{RWF} & \textbf{Grad Norm} & \textbf{Causal} & \textbf{Rel. $L^2$ error} & \textbf{Run time (min)} \\ 
    \midrule
    $\cmark$ & $\cmark$ & $\cmark$ & $\cmark$ & $\cmark$ & $1.02 \times 10^{-2}$ & 9.18 \\ 
    $\xmark$ & $\cmark$ & $\cmark$ & $\cmark$ & $\cmark$ &  $7.37 \times 10^{-1}$ &  8.76 \\ 
    $\cmark$ & $\xmark$ & $\cmark$ & $\cmark$ & $\cmark$ &  $4.29 \times 10^{-1}$ & 7.60 \\ 
    $\cmark$ & $\cmark$ & $\xmark$ & $\cmark$ & $\cmark$ &  $1.31 \times 10^{-2}$ & 9.25 \\ 
    $\cmark$ & $\cmark$ & $\cmark$ & $\xmark$ & $\cmark$ &  $1.13 \times 10^{0}$ &  7.46 \\ 
     $\cmark$ & $\cmark$ & $\cmark$ & $\cmark$ & $\xmark$ &  $1.49 \times 10^{-2}$ &  9.18 \\ 
     $\xmark$ & $\xmark$ & $\xmark$ & $\xmark$ & $\xmark$ &  $9.51 \times 10^{-1}$ &  7.12 \\
    \bottomrule
    \end{tabular}
    \caption{{\em Advection equation:}  Relative $L^2$ error and run time for an ablation study illustrating the impact of disabling individual components of the proposed technique and training pipeline.}
    \label{tab: adv}
\end{table}

\begin{figure}
    \centering
    \includegraphics[width=0.7\textwidth]{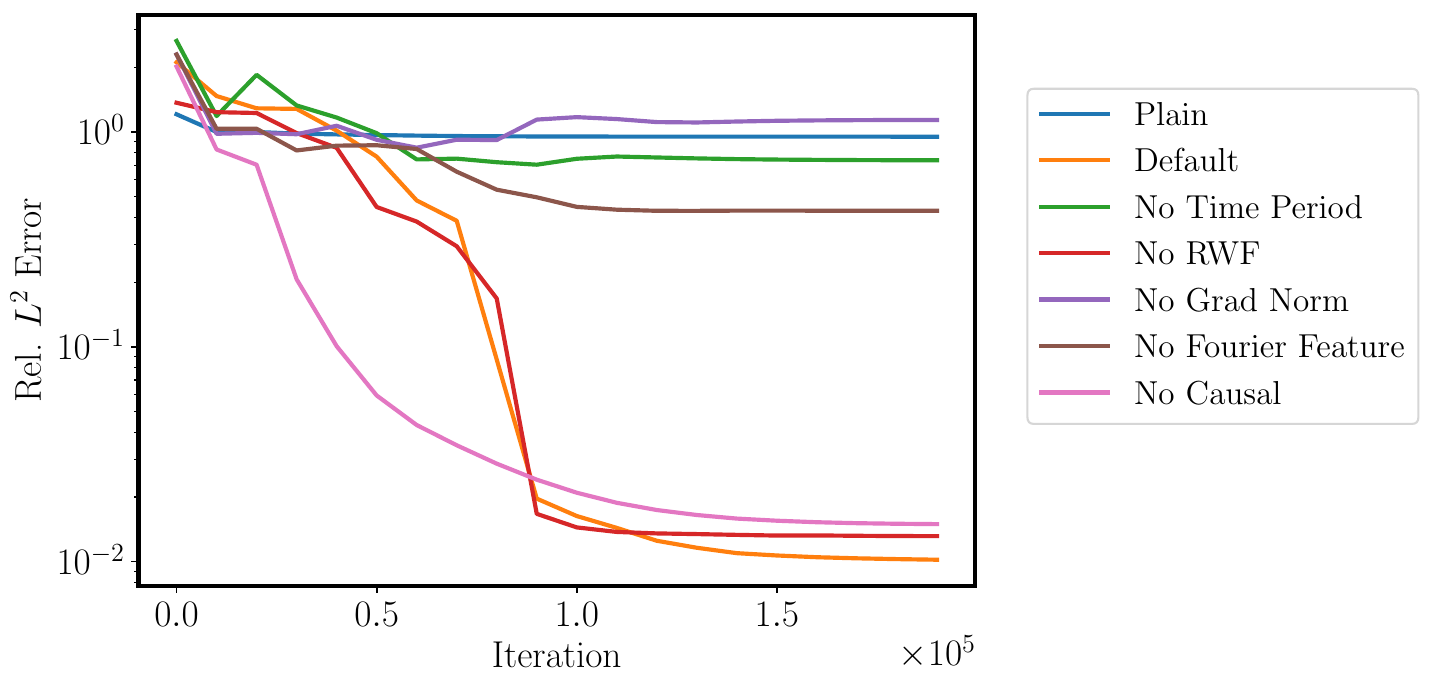}
    \caption{{\em Advection equation:}  Convergence of relative $L^2$ error for the ablation study with different components disabled. 
    Plain: Conventional PINN formulation. Default: PINN model trained imposing time periodicity and using Algorithm \ref{alg: pipline}. No Time Period: PINN model trained using Algorithm \ref{alg: pipline}.
    No RWF: PINN model trained imposing time periodicity and using Algorithm \ref{alg: pipline} without random weight factorization. 
    No Grad Norm: PINN model trained imposing time periodicity and using Algorithm \ref{alg: pipline} without grad norm weighting scheme. 
    No Fourier feature: PINN model trained imposing time periodicity and using Algorithm \ref{alg: pipline} without random Fourier feature embeddings. 
    No Causal: PINN model trained imposing time periodicity and using Algorithm \ref{alg: pipline} without casual weighting.}
    \label{fig: adv_ablation}
\end{figure}

\subsection{Stokes flow}
\label{sec: stokes}

In this example, we explore a specific example of Stokes flow with the aim of emphasizing the importance of non-dimensionalization in PINNs training. Stokes flow is a fluid flow regime where viscous forces outweigh inertial forces, occurring in scenarios such as small particle motion in liquids, fluid flow through porous media, and microorganism locomotion in fluid environments. The governing equation is given by
\begin{align}
    -\nu \Delta \mathbf{u} +\nabla p=0, \\
    \nabla \cdot \mathbf{u} = 0,
\end{align}
where $\mathbf{u} = (u, v)$ defines the velocity and $p$ the pressure, and $\nu$ is the  kinematic viscosity.

\begin{figure}
    \centering
    \includegraphics[width=0.9\textwidth]{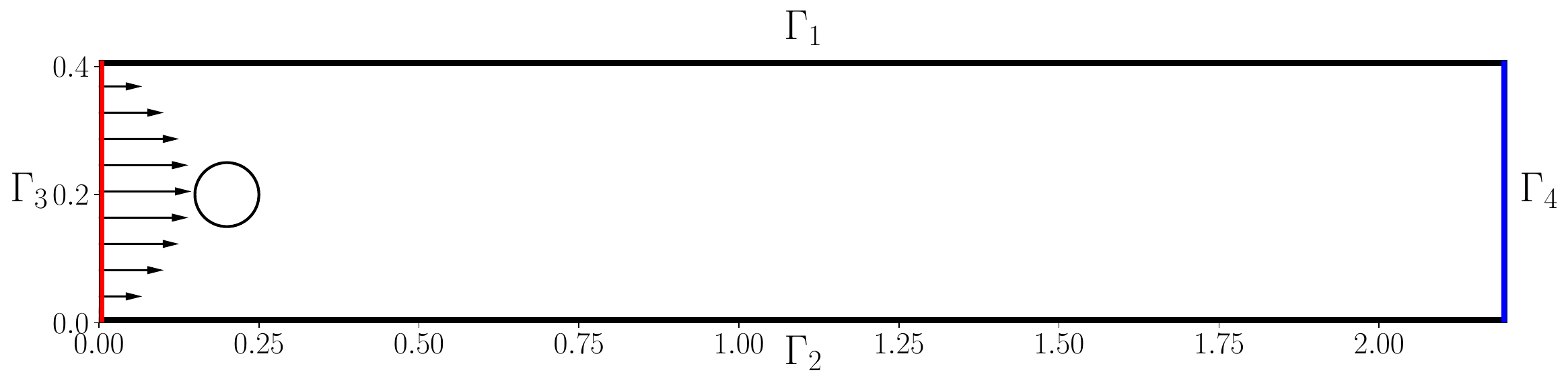}
    \caption{{\em Stokes equation:}  Illustration of the pipe geometry for Stokes flow.}
    \label{fig: stokes_geometry}
\end{figure}

As depicted in Figure \ref{fig: stokes_geometry}, the underlying geometry is a pipe $\Omega=[0,2.2] \times[0,0.41] \backslash B_r(0.2,0.2)$ with a circular cylinder obstacle of radius $r=0.05$.  For the top and bottom walls  $\Gamma_{1}=[0,2.2] \times 0.41$ and $\Gamma_{2}=[0,2.2] \times 0$ as well as the boundary $S=\partial B_r(0.2,0.2)$, we impose the no-slip boundary condition
\begin{align}
    u_{\mid \Gamma_{1}}=u_{\mid \Gamma_{2}}=u_{\mid S}=0.
\end{align}
At the inlet $\Gamma_3=0 \times[0,0.41]$, a parabolic inflow profile is prescribed,
\begin{align}
    \mathbf{u}(0, y)= \mathbf{u}_{\mathrm{in}} = \left(\frac{4 U y(0.41-y)}{0.41^2}, 0\right),
\end{align}
with a maximum velocity $U=0.3$. At the outlet $\Gamma_4=2.2 \times[0,0.41]$, we define the outflow condition
\begin{align}
    \nu \partial_\mathbf{n} \mathbf{u}-p \mathbf{n}=0,
\end{align}
where $\mathbf{n}$ denotes the outer normal vector.

To non-dimensionalize the system, we select the characteristic flow velocity and length as $U^* = 0.2$ and $L^* = 0.1$, respectively. This results in a Reynolds number of
\begin{align}
  \text{Re}=\frac{U^{*} L^*}{\nu}=\frac{0.2 \cdot 0.1}{0.001}= 20.
\end{align}

We can then define the non-dimensionalized variables as
\begin{align}
   \mathbf{x}^* = \frac{\mathbf{x}}{L^*}, \quad  \mathbf{u}^* &=\frac{\mathbf{u}}{U^*}, \quad p^*=\frac{p L^*}{\nu U^*}, \quad \nabla^*=L^* \nabla.
\end{align}
By substituting these scales into the dimensionalized system, we obtain the non-dimensionalized PDE as 
\begin{align}
    -\frac{1}{\text{Re}} \Delta \mathbf{u}^*+\nabla^* p^* & =\mathbf{0} & &\text { in } \Omega^*, \\
    \nabla^* \mathbf{u}^* & =0 & & \text { in } \Omega^*, \\
    \mathbf{u}^* & =0 & & \text { on } \Gamma_{1}^* \cup \Gamma_{2}^* \cup S^*, \\
    \mathbf{u}^* & = \frac{\mathbf{u}_{\mathrm{in}} }{U^*}   & & \text { on } \Gamma_{3}^*, \\
    \frac{1}{\text{Re}} \frac{\partial \mathbf{u}^*}{\partial \mathbf{n}}-p^* \mathbf{n} & =0 & & \text { on } \Gamma_{4}^*,
\end{align}
where $\Omega^*$, $S^*$ and $\{\Gamma_i\}_{i=1}^4$ denote the non-dimensionalized domains, respectively.

\begin{figure}
    \centering
    \includegraphics[width=0.9\textwidth]{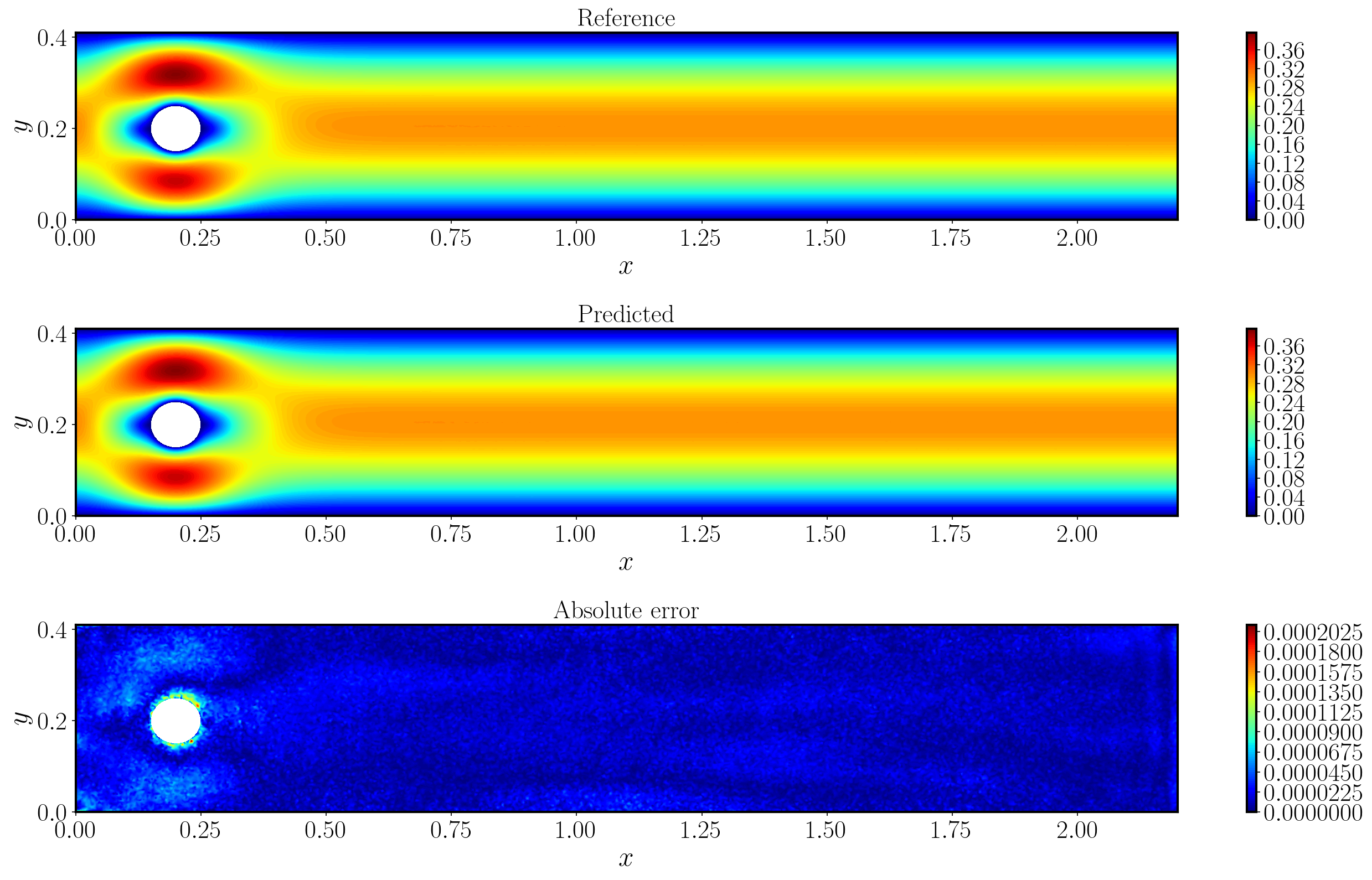}
    \caption{{\em Stokes equation:}  Comparison of the best prediction against the reference solution obtained from the hyper-parameter sweep. The resulting relative $L^2$ error is $8.04 \times 10^{-5}$. The hyper-parameter configuration can be found in Table \ref{tab: stokes_config}.}
    \label{fig: stokes_U}
\end{figure}

To perform an ablation study for Algorithm \ref{alg: pipline}, we employ an MLP with 4 hidden layers, 128 neurons per hidden layer, and GeLU activation functions and train each model for $10^5$ iterations of gradient descent using the Adam optimizer.  The results are summarized in Table \ref{tab: Stokes}, and strongly indicate the positive impact of  all proposed components on model performance; disabling any one component leads to worse predictive accuracy. In particular, comparing the performance of the configurations with non-dimensionalization enabled (1st row) to the ones with non-dimensionalization disabled (5th rows), we observe a substantial increase in the relative $L^2$ error when non-dimensionalization is removed. This observation highlights the  importance of non-dimensionalization in achieving  optimal performance for solving the Stokes equation.  Moreover, as evidenced by the 3rd and 4th rows of the table, models trained without Fourier features and RWF fail to capture the correct solution, thus implying their essential contribution to the overall model performance. Lastly, we present the results of a fine-tuned PINN model in Figure \ref{fig: stokes_U}, which exhibits excellent agreement with the reference solution and achieves a relative $L^2$ error of $8.04 \times 10^{-5}$.  The detailed hyper-parameter configuration and the loss convergence are respectively shown in Table \ref{tab: stokes_config} and Figure \ref{fig: stokes_loss_weight}.

\begin{table}[]
    \centering
    \renewcommand{\arraystretch}{1.2}
    \begin{tabular}{ccccccc}
    \toprule
    \multicolumn{4}{c}{\textbf{Ablation Settings}} & \multicolumn{2}{c}{\textbf{Performance}} \\ 
    \cmidrule(lr){1-4} \cmidrule(lr){5-6}
    \textbf{Fourier Feature} & \textbf{RWF} & \textbf{Grad Norm} & \textbf{Non-dimensionalization} & \textbf{Rel. $L^2$ error} & \textbf{Run time (min) } \\ 
    \midrule
    $\cmark$ & $\cmark$ & $\cmark$ & $\cmark$ & $5.41 \times 10^{-4}$ & 9.51  \\ 
    $\xmark$ & $\cmark$ & $\cmark$ & $\cmark$ &  $9.56 \times 10^{-1}$ &  7.93 \\ 
    $\cmark$ & $\xmark$ & $\cmark$ & $\cmark$ &  $9.86 \times 10^{-1}$ &  9.58 \\ 
    $\cmark$ & $\cmark$ & $\xmark$ & $\cmark$ &  $1.01 \times 10^{-2}$ &  8.63 \\ 
    $\cmark$ & $\cmark$ & $\cmark$ & $\xmark$ &  $9.74 \times 10^{-1}$ &  9.58 \\ 
    $\xmark$ & $\xmark$ & $\xmark$ & $\xmark$ &  $9.21 \times 10^{-1}$ &  7.95 \\ 
    \bottomrule
    \end{tabular}
    \caption{{\em Stokes equation:}   Relative $L^2$ error and run time for an ablation study illustrating the impact of disabling non-dimensionalization and individual components of the proposed training pipeline. The error is measured against  the norm of flow velocity $\|\mathbf{u}\|_2 = \sqrt{u^2 + v^2}$. }
    \label{tab: Stokes}
\end{table}

\begin{figure}
    \centering
    \includegraphics[width=0.7\textwidth]{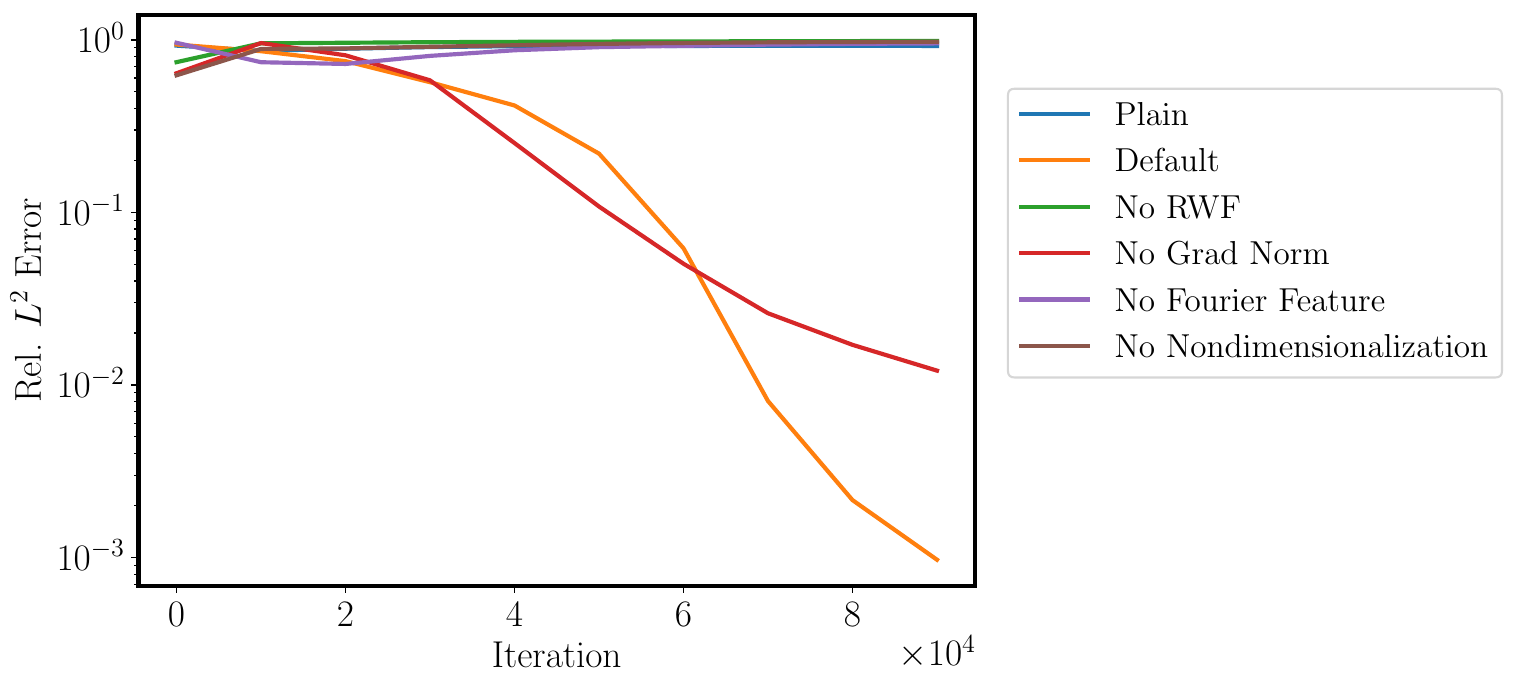}
    \caption{{\em Stokes equation:}  Convergence of relative $L^2$ error for the ablation study with different components disabled. }
    \label{fig: stokes_ablation}
\end{figure}

\subsection{Kuramoto–Sivashinsky equation}


In this example, we aim to demonstrate the potential of PINNs in simulating chaotic dynamics and highlight the necessity of adopting a time-marching strategy in scenarios where high predictive accuracy is needed. To this end, we consider  the Kuramoto–Sivashinsky equation, which exhibits a wealth of spatially and temporally nontrivial dynamical behavior, and has served as a model example in efforts to understand and predict the complex dynamical behavior associated with a variety of physical systems. The equation takes the form
\begin{align}
    \label{eq: ks}
    u_{t}+ \alpha u u_x + \beta  u_{x x}+ \gamma u_{x x x x}=0, \quad t \in [0, 1], \ x \in [0, 2 \pi],
\end{align}
subject to periodic boundary conditions and an initial condition 
\begin{align}
    u(0, x) = u_0(x).
\end{align}
Specifically, we take $\alpha = 100/16, \beta=100/16^2,  \gamma=100/16^4$ and $u_0(x) = \cos(x)(1 + \sin(x))$. 

\begin{figure}
    \centering
    \includegraphics[width=0.9\textwidth]{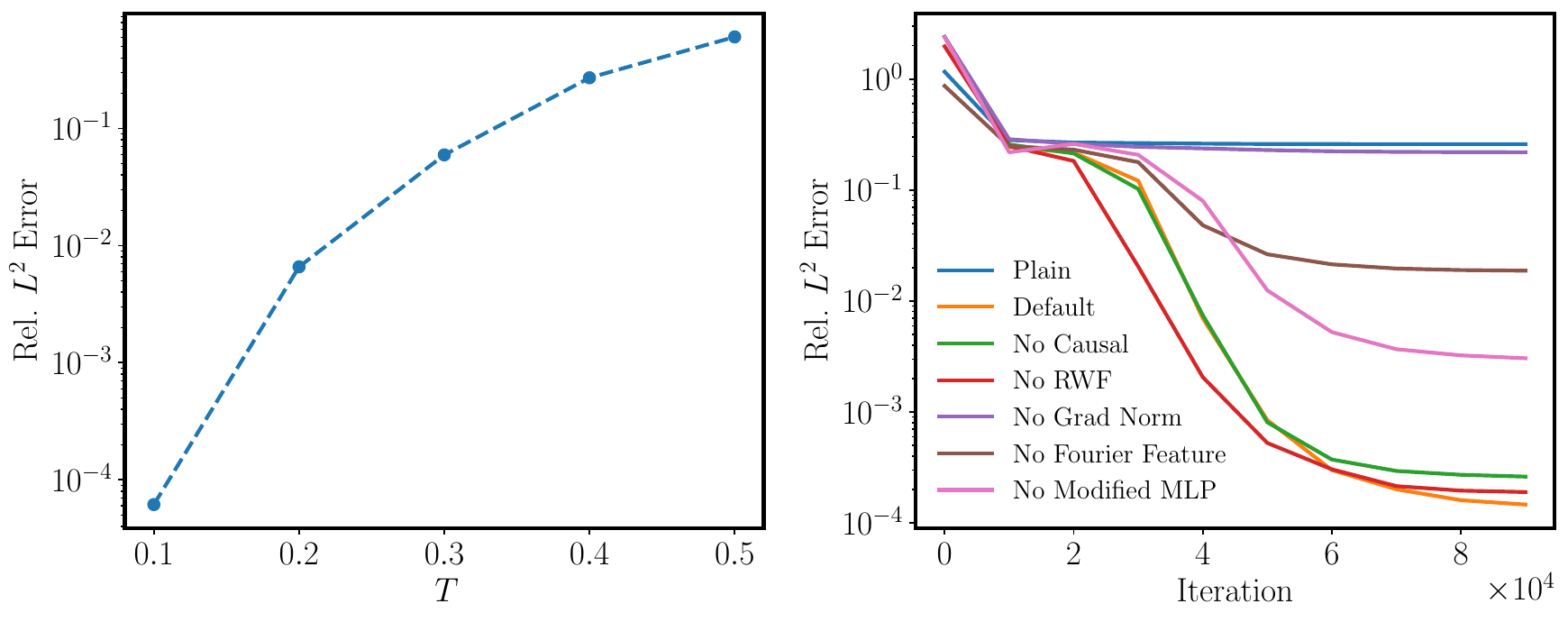}
    \caption{{\em Kuramoto–Sivashinsky equation:} {\em Left:}  Relative $L^2$ errors from one-shot PINN training for different system final time $T$ under the same hyper-parameter setting.  {\em Right:} Convergence of relative $L^2$ error for the ablation study with different components disabled. }
    \label{fig: ks_ablation}
\end{figure}

Based on our experience, it appears highly challenging to conduct long-term integration of this PDE system via a single-shot training of PINNs. This could potentially be attributed to the inherently chaotic nature of the system and the insufficient accuracy of PINNs predictions. To illustrate this point, we train a PINN to simulate the dynamical system up to different final time $T$ without time-marching, while keeping the same hyper-parameter settings.  As shown in the left panel of Figure \ref{fig: ks_ablation}, we can see that the resulting relative $L^2$ error drastically increases for larger temporal domains and eventually leads to a failure in correctly capturing the PDE solution. This illustrates the necessity for applying time-marching in order to mitigate the difficulties of approximation and optimization, thus leading to more accurate results. However, we must emphasize that the computational cost of time-marching is considerably larger than one-shot learning as one needs to train multiple PINN models sequentially. It would be interesting to explore the acceleration of this training process in the future work.

\begin{figure}
    \centering
    \includegraphics[width=0.9\textwidth]{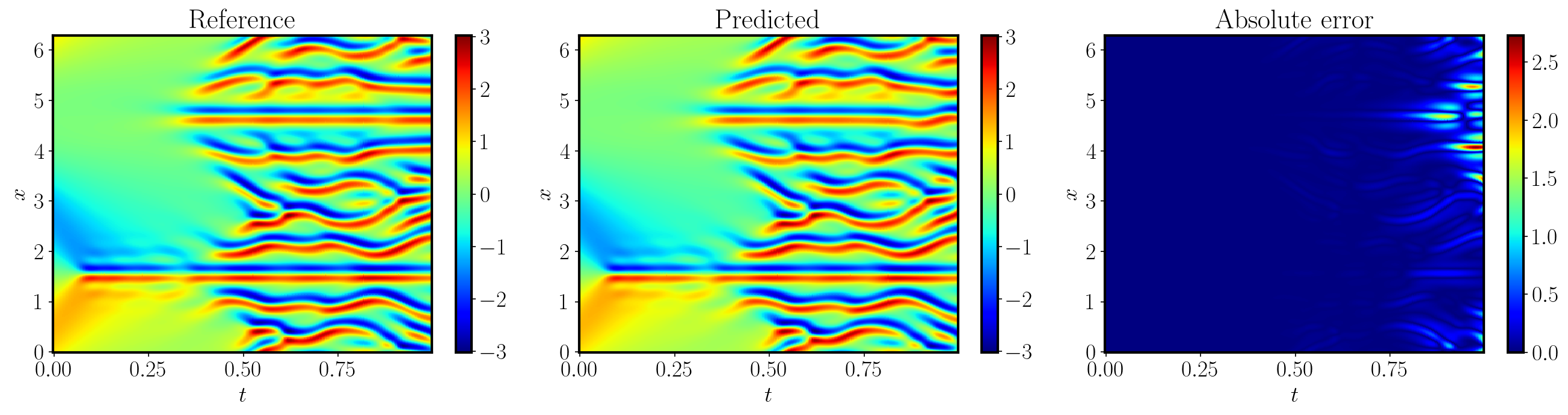}
    \caption{{\em Kuramoto–Sivashinsky equation:}  Comparison of the best prediction against the reference solution. The relative $L^2$ error of the spatial temporal predicted solution is $1.61 \times 10^{-1}$. Note that the the majority of this error is attributed to last few time steps.}
    \label{fig: ks_pred}
\end{figure}

Moreover, we present an ablation study on Algorithm \ref{alg: pipline} and summarize our results in Table \ref{tab: ks}. It can be concluded that all proposed components positively contribute to the overall model performance, and removing of any one of them results in increased errors. Notably, the 
use of modified MLP greatly enhances the predictive accuracy, reflected in the substantial error reduction from $2.98 \times 10^{-3}$ to $1.42 \times 10^{-4}$. From our experience, modified MLPs typically outperforms plain MLPs, especially for tackling non-linear PDE systems. Furthermore, the predicted solution obtained from our best model is visualized in Figure \ref{fig: ks_pred}, which is in a good agreement with the ground truth. Nevertheless,  some discrepancies can be observed near $t=1$, which may be due to the error accumulation and the inherent nature of chaos. More details of implementation and training are provided in Table \ref{tab: ks_config} and Figure \ref{fig: ks_loss_weight}.

\begin{table}[]
    \centering
    \renewcommand{\arraystretch}{1.2}
    \begin{tabular}{ccccccc}
    \toprule
    \multicolumn{5}{c}{\textbf{Ablation Settings}} & \multicolumn{2}{c}{\textbf{Performance}} \\ 
    \cmidrule(lr){1-5} \cmidrule(lr){6-7}
    \textbf{Modified MLP} & \textbf{Fourier Feature} & \textbf{RWF} & \textbf{Grad Norm} & \textbf{Causal} & \textbf{Rel. $L^2$ error} & \textbf{Run time (min)} \\ 
    \midrule
    $\cmark$ & $\cmark$ & $\cmark$ & $\cmark$ & $\cmark$ & $\mathbf{1.42 \times 10^{-4}}$ & 13.33 \\ 
    $\xmark$ & $\cmark$ & $\cmark$ & $\cmark$ & $\cmark$ &  $2.98 \times 10^{-3}$ &  6.21 \\ 
    $\cmark$ & $\xmark$ & $\cmark$ & $\cmark$ & $\cmark$ &  $1.86 \times 10^{-2}$ & 7.60 \\ 
    $\cmark$ & $\cmark$ & $\xmark$ & $\cmark$ & $\cmark$ &  $1.86 \times 10^{-4}$ & 14.11 \\ 
    $\cmark$ & $\cmark$ & $\cmark$ & $\xmark$ & $\cmark$ &  $2.19 \times 10^{-1}$ &  14.11 \\ 
     $\cmark$ & $\cmark$ & $\cmark$ & $\cmark$ & $\xmark$ &  $2.58 \times 10^{-4}$ &  9.18 \\ 
     $\xmark$ & $\xmark$ & $\xmark$ & $\xmark$ & $\xmark$ &  $2.59 \times 10^{-1}$ &  7.12 \\
    \bottomrule
    \end{tabular}
    \caption{{\em Kuramoto–Sivashinsky equation:}  Relative $L^2$ error and run time for an ablation study illustrating the impact of disabling individual components of the proposed technique and training pipeline. }
    \label{tab: ks}
\end{table}

\subsection{Lid-driven cavity flow}

\label{sec: ldc}

In this example, we consider a classical benchmark problem in computational fluid dynamics, describing the motion of an incompressible fluid in a two-dimensional square cavity. The system is governed by the incompressible Navier–Stokes equations written in a non-dimensional form
\begin{align}
    \mathbf{u} \cdot \nabla \mathbf{u}+\nabla p-\frac{1}{R e} \Delta \mathbf{u}&=0, \quad  (x,y) \in (0,1)^2, \\
    \nabla \cdot \mathbf{u}&=0, \quad  (x,y) \in (0,1)^2,
\end{align}
where $\mathbf{u} = (u, v)$ denotes the velocity in $x$ and $y$ directions, respectively, and $p$ is the scalar pressure field. We assume $\mathbf{u}=(1, 0)$ on the top lid of the cavity, and a non-slip boundary condition on the other three walls. We are interested in the velocity and pressure distribution for a Reynolds number of $3200$.

\begin{figure}
    \centering
    \includegraphics[width=0.4\textwidth]{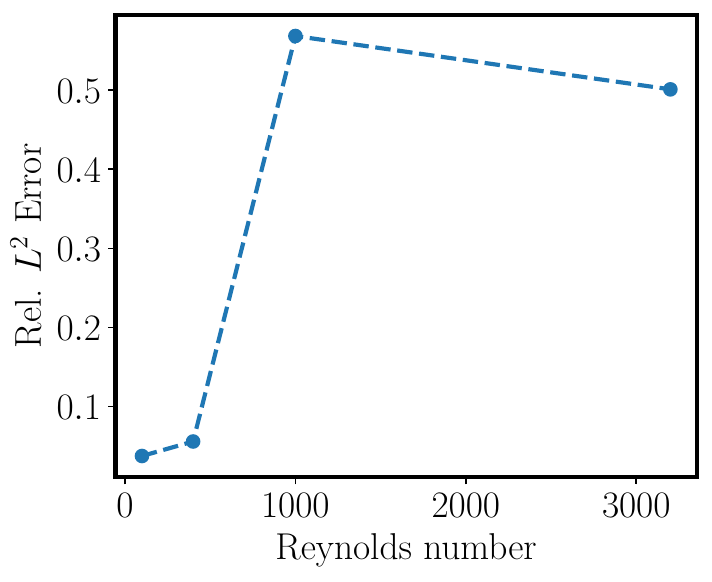}
    \caption{{\em Lid-driven cavity:} Relative $L^2$ error of training PINNs with Algorithm \ref{alg: pipline} at different  Reynolds numbers $Re \in [100, 400, 1000, 3200]$. }
    \label{fig: ldc_l2_error}
\end{figure}

\begin{figure}
    \centering
    \includegraphics[width=0.9\textwidth]{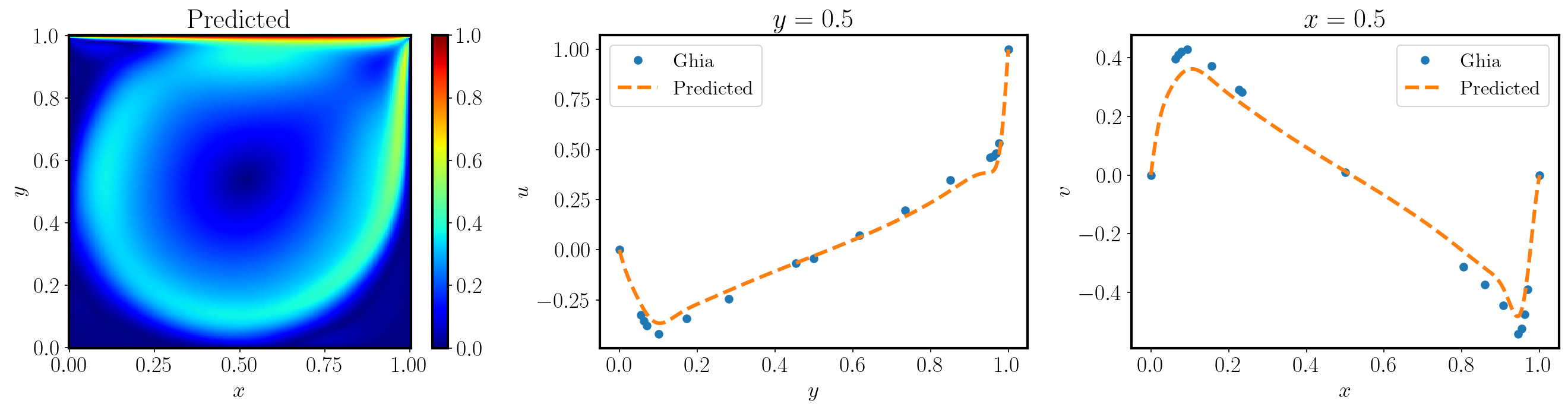}
    \caption{{\em Lid-driven cavity (Re=3200):} {\em Left:} Predicted velocity of the fine-tuned model. {\em Right:} Comparison of the predicted velocity profiles on the vertical and horizontal center-lines against Ghia {\em et al.} \cite{GHIA1982387}. The resulting relative $L^2$ error against the reference solution is $1.58 \times 10^{-1}$.}
    \label{fig: ldc_pred}
\end{figure}

In our experience, when trained directly at a high Reynolds number, PINNs tend to be unstable and susceptible of converging to erroneous solutions. 
This observation is verified by the left panel of Figure \ref{fig: ldc_l2_error}, where we plot the  relative $L^2$ errors  from training PINNs with Algorithm \ref{alg: pipline} at varying Reynolds numbers under the same hyper-parameter settings.  Our results demonstrate that PINNs struggle to yield accurate solutions for Reynolds numbers greater than $1,000$. To improve  this result, one effective approach is to start the training of PINNs with a lower initial Reynolds number, and gradually increase the Reynolds numbers during training. By this way, the model parameters obtained from the training with smaller Reynolds numbers  serve as a good initialization  when training for higher Reynolds numbers. To demonstrate this, we select an increasing sequence of Reynolds numbers $[100, 400, 1000, 3200]$ and train PINNs with Algorithm \ref{alg: pipline} for $5 \times 10^4, 5 \times 10^4, 1 \times 10^5, 5 \times 10^5$ iterations, respectively. The detailed hyper-parameter configuration  is summarized in Table \ref{tab: ldc_config}. As shown in Figure \ref{fig: ldc_pred}, our predicted velocity field agrees well with the reference results of Ghia {\it et al.} \cite{GHIA1982387}, yielding a relative $L^2$ error of $1.58 \times 10^{-1}$ against the reference solution.

\subsection{Navier–Stokes flow in a torus}

\label{sec: ns_tori}

As the second to last example, our goal is to showcase the capability of PINNs in simulating incompressible Navier–Stokes flow using the velocity-vorticity formulation. The equation is given by
\begin{align}
w_t +\mathbf{u} \cdot \nabla w &= \frac{1}{\text{Re}} \Delta w,   \quad \text{ in }  [0, T] \times \Omega,  \\
\nabla \cdot \mathbf{u}  &=0,  \quad \text{ in }  [0, T] \times \Omega, \\
w(0, x, y) &=w_{0}(x, y),   \quad \text{ in }  \Omega.
\end{align}
Here, $\mathbf{}{u} = (u, v)$ represents the flow velocity field, $w = \nabla \times \mathbf{u}$ denotes the vorticity, and $\text{Re}$ denotes the Reynolds number. For this example, we define $\Omega = [0, 2\pi]^2$ and set $\text{Re}$ as 100. 

\begin{figure}
    \centering
    \includegraphics[width=0.9\textwidth]{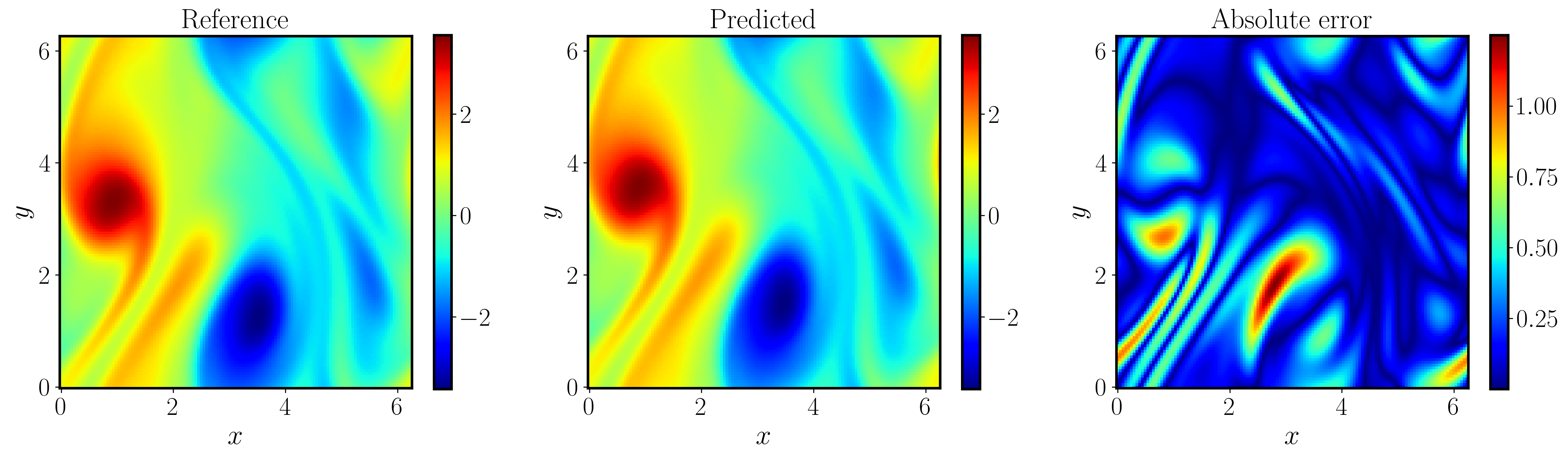}
    \caption{{\em Navier-Stokes flow in a torus:}  Comparison of the best prediction against the reference solution at the last time step. The animation is provided in \url{https://github.com/PredictiveIntelligenceLab/jaxpi}. }
    \label{fig: ns_w}
\end{figure}

As the validation and effectiveness of the proposed PINN algorithm have been rigorously proven in prior examples, our focus is on simulating the vorticity evolution up to $T = 10$ using PINNs.  To this end, we split the temporal domain into 5 intervals and employ a time-marching strategy. For each interval, we use a PINN model with a modified MLP (4 hidden layers, 256 neurons per hidden layer, Tanh activations) and train it using Algorithm  \ref{alg: pipline} for $10^5$ iterations of gradient descent with Adam optimizer. The results of this simulation are summarized in Figure \ref{fig: ns_w}, which  provides a visual comparison of the reference and predicted vorticity at $T=10$. While a slight misalignment between the two can be observed, the model  prediction is in good agreement with the corresponding numerical estimations. This demonstrates the capability of PINNs to closely match the reference solution, emphasizing its effectiveness in simulating vortical fluid flows.

\subsection{Navier–Stokes flow around a cylinder}

In our last example, we investigate  a classical benchmark  in computational fluid dynamics, describing the behaviour of a transient fluid in a pipe with a circular obstacle.
Previous research by Chuang {\em et al.} \cite{chuang2022experience}  reported that PINNs  act as a steady-flow solver, and fail to capture the phenomenon of vortex shedding. Here we challenge these findings and demonstrate that, if properly used, PINNs can successfully simulate the development of vortex shedding in this scenario. 

Specifically, we consider a fluid with a density of  $\rho=1.0$ and  describe its behavior using the time-dependent incompressible Navier-Stokes equations
\begin{align}
       \mathbf{u}_t + \mathbf{u} \nabla \mathbf{u} + \nabla p - \nu \mathbf{u} = 0, \\
    \nabla \cdot \mathbf{u} = 0,
\end{align}
with $\mathbf{u}=(u, v)$ defining the velocity field and $p$ the pressure. The kinematic viscosity is taken as $\nu =0.001$.

The underlying geometry is identical to Figure \ref{fig: stokes_geometry} and the boundary conditions are the same as the Stokes flow  example discussed in section \ref{sec: stokes}.
However, we introduce a parabolic inflow profile with a maximum velocity of $U=1.5$. As a result, we have characteristic flow velocity and length values of $U=1.0$ and $L=0.1$, respectively, and a Reynolds number of $Re=100$.

We begin by normalizing the PDE system as follows:
\begin{align}
   \mathbf{x}^* = \frac{\mathbf{x}}{L^*}, \quad   t^* = 
\frac{L^*}{U^*}, \quad \mathbf{u}^* &=\frac{\mathbf{u}}{U^*}, \quad p^*=\frac{p L^*}{\nu U^*}, \quad \nabla^*=L^* \nabla.
\end{align}
This leads us to the non-dimensionalized equations:
\begin{align}
       \mathbf{u}^*_t + \mathbf{u}^* \nabla^* \mathbf{u} + \nabla p^* - \frac{1}{Re}\mathbf{u}^* = 0, \\
    \nabla^* \cdot \mathbf{u}^* = 0.
\end{align}

To obtain a proper initial condition for PINNs, we start with a zero solution and run a numerical simulation for $4$ seconds at a very coarse spatial and temporal resolution. We then use the last time-step as our initial condition for the PINNs simulation.

\begin{figure}
    \centering
    \includegraphics[width=0.9\textwidth]{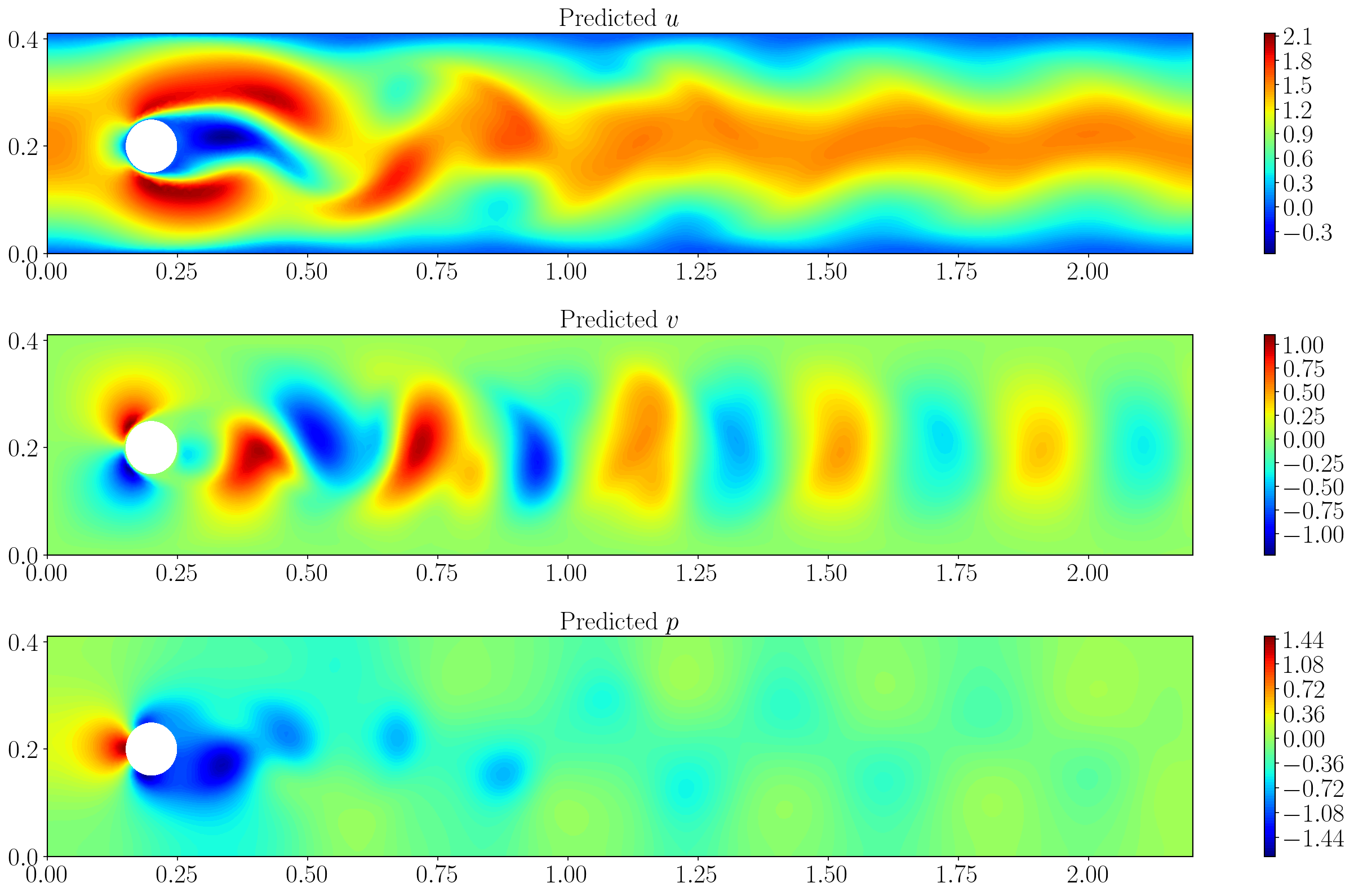}
    \caption{{\em Navier-Stokes flow around cylinder:} Predicted velocity field and pressure at $T=1$. the last time step. The animation is provided in \url{https://github.com/PredictiveIntelligenceLab/jaxpi}.}
    \label{fig: ns_cylinder_pred}
\end{figure}

Using a time-marching strategy, we partition the temporal domain $[0, 10]$ into 10 individual time windows. For each window, a modified MLP is employed as our model backbone. PINN training runs for $2 \times 10^5$ iterations per window following Algorithm \ref{alg: pipline}. Key hyper-parameters are detailed in Table \ref{tab: ns_cylinder_config}. It deserves mentioning that there are more than 10 terms in the total loss and thus it is practically infeasible to manually adjust the weights of each loss. The predicted velocity and pressure field at $T=10$ are plotted in Figure \ref{fig: ns_cylinder_pred}. For this benchmark, we do not report the test error against the numerical solution, as the start time of vortex shedding in numerical solvers fluctuates based on the underlying discretizations. To the best of our knowledge, our work presents the first empirical evidence of a PINN model being able to capture the phenomenon of vortex shedding. This finding opens up new avenues for further research and application of PINNs in the field of computational fluid dynamics.

\section{Conclusions}

In this work, we introduce a comprehensive training pipeline for physics-informed neural networks, addressing various training pathologies such as spectral bias, imbalanced losses, and causality violation. 
Our pipeline seamlessly integrates  essential techniques, including equation non-dimensionalization, Fourier feature embeddings, loss weighting schemes and  causal training strategies. Moreover, we explore additional techniques such as a modified MLP architecture, random weight factorization and curriculum training, which can further improve the training stability and model performance. 
By sharing our empirical findings, we also provide insights into selecting appropriate hyper-parameters associated with network architectures and learning rate schedules in conjunction with the aforementioned algorithms.
To demonstrate the effectiveness of the proposed training pipeline, we perform thorough ablation studies on a collection of  benchmarks which PINNs often struggle with, and showcase the state-of-the-art results, which we believe should serve as a strong baseline for future studies.
By establishing these benchmarks, we hope that  our contribution will serve as a cornerstone for more fair and systematic comparisons in the development and adoption of PINN-based methodologies, ultimately propelling PINN research towards more effective and reliable solutions in  computational science and engineering.





\section*{Acknowledgments}
We would like to acknowledge support from the US Department of Energy under the Advanced Scientific Computing Research program (grant DE-SC0019116), the US Air Force (grant AFOSR FA9550-20-1-0060), and US Department of Energy/Advanced Research Projects Agency (grant DE-AR0001201). We also thank the developers of the software that enabled our research, including JAX \cite{jax2018github}, JAX-CFD\cite{Kochkov2021}, Matplotlib \cite{hunter2007matplotlib}, and NumPy \cite{harris2020array}.

\bibliographystyle{unsrt}  
\bibliography{references}

\appendix

\section{Spectral Bias through the lens of the Neural Tangent Kernel}
\label{appendix: ntk}

We investigate spectral bias \cite{rahaman2019spectral, cao2019towards, ronen2019convergence} in the training behavior of deep fully-connected networks through the lens of Neural Tangent Kernel(NTK) theory.
Let $f_{\mathbf{\theta}}(\mathbf{x})$ be a scalar-valued fully-connected neural network. Given a training data-set $\{\mathbf{X}_{\text{train}}, \mathbf{Y}_{\text{train}}\}$, where $\mathbf{X}_{\text{train}} = (\mathbf{x}_i)_{i=1}^N$ are inputs and $\mathbf{Y}_{\text{train}} = (y_i)_{i=1}^N$ are the corresponding labels.  We consider a network trained by minimizing the mean square loss 
\begin{align}
    \mathcal{L}(\mathbf{\theta}) = \frac{1}{N}\sum_{i=1}^N |f_{\mathbf{\theta}}(\mathbf{x}_i) - y_i|^2.
\end{align}
Following the derivation of Jacot {\em et al.} \cite{jacot2018neural, arora2019exact}, we can define the resulting neural tangent kernel  $\mathbf{K}$, whose $ij$-th entry is given by
\begin{align}
    \label{eq: NTK}
   \mathbf{K}_{ij} = \mathbf{K}(\mathbf{x}_i, \mathbf{x}_j) =  \left\langle\frac{\partial f_{\mathbf{\theta}}(\mathbf{x}_i)}{\partial \boldsymbol{\theta}}, \frac{\partial f_{\mathbf{\theta}}\left(\mathbf{x}_j\right)}{\partial \boldsymbol{\theta}}\right\rangle.
\end{align}
The NTK theory shows that, under gradient descent dynamics with an infinitesimally small learning rate (gradient flow),
the kernel $\mathbf{K}$ converges to a deterministic kernel $\mathbf{K}^*$ and does not change during training as the width of the network grows to infinity. 

Furthermore, under the asymptotic conditions stated in Lee {\em et al.} \cite{lee2019wide}, we may derive that
\begin{align}
    \frac{d f_{\mathbf{\theta}(\tau)}(\mathbf{X}_{\text{train}})}{ d \tau} \approx - \mathbf{K} \cdot \left(f_{\mathbf{\theta}(\tau)}(\mathbf{X}_{\text{train}}) - \mathbf{Y}_{\text{train}} \right),
\end{align}
where $\mathbf{\theta}(\tau)$ denotes the parameters of the network at iteration $\tau$ and $f_{\mathbf{\theta}(\tau)}(\mathbf{X}_{\text{train}} ) = (f_{\mathbf{\theta}(\tau)}(\mathbf{x}_i)_{i=1}^N$.  Then, it directly follows that
\begin{align}
     f_{\mathbf{\theta}(\tau)}\left(\mathbf{X}_{\text{train}}) \approx (I -  e^{-\mathbf{K}\tau} \right) \cdot \mathbf{Y}_{\text{train}}.
\end{align}
Since the kernel $\mathbf{K}$ is positive semi-definite, we can take its spectral decomposition $\mathbf{K} = \mathbf{Q}^T\mathbf{\Lambda} \mathbf{Q}$, where $\mathbf{Q}$ is an orthogonal matrix whose $i$-th column is the eigenvector $\mathbf{q}_i$ of $\mathbf{K}$ and $\mathbf{\Lambda}$ is a diagonal matrix whose diagonal entries $\lambda_i$ are the corresponding eigenvalues. 
Since $e^{-\mathbf{K}t} = \mathbf{Q} e^{-\mathbf{\Lambda} \tau} \mathbf{Q}^T$, we have 
\begin{align}
    \mathbf{Q}^T \left(f_{ \mathbf{\theta}(\tau)  }(\mathbf{X}_{\text{train}}) -       \mathbf{Y}_{\text{train}} \right) = - e^{\mathbf{\Lambda} \tau} \mathbf{Q}^T \mathbf{Y_{\text{train}}},
\end{align}
which implies
\begin{align}
      \begin{bmatrix}
        \mathbf{q}_1^T \\
        \mathbf{q}_2^T \\
        \vdots \\
        \mathbf{q}_N^T
    \end{bmatrix}  ( f_{ \mathbf{\theta}(\tau)  }(\mathbf{X}_{\text{train}}) -       \mathbf{Y}_{\text{train}}) &=  \begin{bmatrix}
        e^{-\lambda_1 \tau} & & & \\
        & e^{-\lambda_2 \tau} & & \\
                    &    &   \ddots &\\
                    &    &  & e^{-\lambda_N \tau}
    \end{bmatrix}
    \begin{bmatrix}
        \mathbf{q}_1^T \\
        \mathbf{q}_2^T \\
        \vdots \\
        \mathbf{q}_N^T
    \end{bmatrix}
    \mathbf{Y_{\text{train}}}.
\end{align}
The above equation shows that the convergence rate of $\mathbf{q}_i^T  ( f_{ \mathbf{\theta}(\tau) }(\mathbf{X}_{\text{train}}) - \mathbf{Y}_{\text{train}}) $ is determined by the $i$-th eigenvalue $\lambda_i$. 
Moreover, we can decompose the training error into the eigen-space of the NTK as
\begin{align}
f_{\mathbf{\theta}(\tau)}(\mathbf{X}_{\text{train}}) -       \mathbf{Y}_{\text{train}}  &= \sum_{i=1}^N (f_{\mathbf{\theta}(\tau)}(\mathbf{X}_{\text{train}}) -       \mathbf{Y}_{\text{train}}   , \mathbf{q}_i) \mathbf{q}_i \\
&=  \sum_{i=1}^N \mathbf{q}_i^T \left(f_{\mathbf{\theta}(\tau)}(\mathbf{X}_{\text{train}}) -       \mathbf{Y}_{\text{train}} \right)\mathbf{q}_i \\
&= \sum_{i=1}^N \left( e^{-\lambda_i \tau} \mathbf{q}_i^T \mathbf{Y}_{\text{train}} \right)\mathbf{q}_i.
\end{align}
Clearly, the network is biased to first learn the target function along the eigen-directions of neural tangent kernel with larger eigenvalues, and then the remaining components corresponding to smaller eigenvalues. Cao {\em et al.} \cite{cao2019towards} provide a detailed analysis of the convergence rate of these components. 
For conventional fully-connected neural networks, the eigenvalues of the NTK shrink monotonically as the frequency of the corresponding eigenfunctions increases, yielding a significantly lower convergence rate for high frequency components of the target function \cite{rahaman2019spectral, ronen2019convergence}.
This indeed reveals the so-called ``spectral bias'' \cite{rahaman2019spectral} pathology of deep neural networks. More importantly, we may conclude that the eigen-space of the neural tangent kernel characterizes the learnability of a target function by a neural network.   

\section{Random Weight Factorization}
\label{appendix: rwf}

In this section, we  provide an intuitive understanding and some theoretical explanations of  random weight factorization. For additional numerical validations of RWF, please see \cite{wang2022random}.

To provide an intuitive understanding, let us consider the simplest setting of a one-parameter loss function $\ell(w)$. In this case, the weight factorization can be simplified to $w = s \cdot v$ with two scalars $s$ and $v$. It is important to note that for any given $w \neq 0$, there exist infinitely many pairs $(s,v)$ such that $w=s \cdot v$. These pairs form a family of hyperbolas in the $sv$-plane, with one for each choice of signs for both $s$ and $v$. Consequently, the loss function in the $sv$-plane remains constant along these hyperbolas.

\begin{figure}
    \centering
    \includegraphics[width=0.75\textwidth]{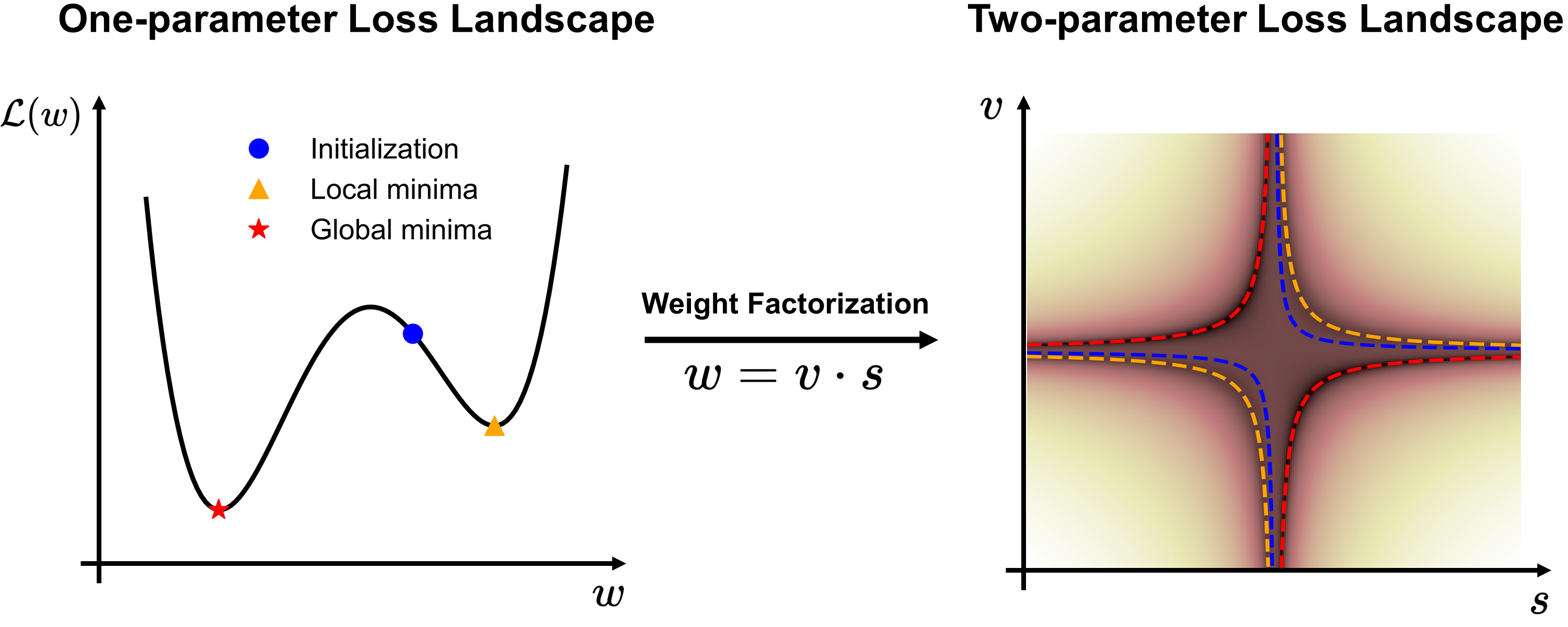}
     \caption{{ Weight factorization transforms loss landscapes and shortens the distance to minima.} }
     \label{fig: loss_landscape}
\end{figure}

Figure \ref{fig: loss_landscape} gives a visual illustration of the difference between the original loss landscape as a function of $w$ versus the loss landscape in the factorized $sv$-plane.  In the left panel, we plot the original loss function as well as an initial parameter point, the local minimum, and the global minimum. The right panel shows how in the factorized parameter space, each of these three points corresponds to two hyperbolas in the $sv$-plane. Note how the distance between the initialization and the global minima is reduced from the left to the right panel upon an appropriate choice of factorization. The key observation is that the distance between factorizations representing the initial parameter and the global minimum becomes arbitrarily small in the $sv$-plane for larger values of $s$.  Indeed, we can prove that this holds for any general loss function in arbitrary parameter dimensions. Further details can be found in \cite{wang2021eigenvector}.

\begin{theorem}
Suppose that $\mathcal{L}(\mathbf{\theta})$ is the associated loss function of a neural network defined in \eqref{eq: mlp_1} and \eqref{eq: mlp_2}. For a given $\mathbf{\theta}$, we define $U_{\mathbf{\theta}}$ as the set containing all possible weight factorizations 
\begin{align}
    U_{\mathbf{\theta}} = \left\{ (\mathbf{s}^{(l)}, \mathbf{V}^{(l)})_{l=1}^{L+1}  : \mathrm{diag}(\mathbf{s}^{(l)}) \cdot  \mathbf{V}^{(l)}   = \mathbf{W}^{(l)}, \quad l=1, \dots, L+1  \right\}.
\end{align}
Then for any $\mathbf{\theta}, \mathbf{\theta}'$, we have
\begin{align}
    \text{dist}(U_{\mathbf{\theta}}, U_{\mathbf{\theta}'}) := \min_{\mathbf{x} \in U_{\mathbf{\theta}}, \mathbf{y} \in U_{\mathbf{\theta}'}} \|\mathbf{x} - \mathbf{y}\|   = 0.
\end{align}
\end{theorem}
\label{appendix: thm1}

\begin{proof}
Starting from any fixed network parameters  $\mathbf{\theta} = \{\mathbf{W}^{(l)}, \mathbf{b}^{l}\}_{l=1}^{L+1}$, we consider the following weight factorization
\begin{align}
    \text{diag}( \mathbf{s}^{(l)})  \cdot \mathbf{V}^{(l)} = \mathbf{W}^{(l)}, \quad l=1, 2, \dots, L+1.
\end{align}

Next, consider the set of all possible weight factorizations associated with the initialization $\mathbf{\theta}$ as
\begin{align}
    U_{\mathbf{\theta}} = \left\{ (\mathbf{s}^{(l)}, \mathbf{V}^{(l)})_{l=1}^{L+1}  : \mathrm{diag}(\mathbf{s}^{(l)}) \cdot  \mathbf{V}^{(l)}   = \mathbf{W}^{(l)}, \quad l=1, \dots, L+1  \right\}.
\end{align}
Let us now define $U_0$ in the factorized parameter space by
\begin{align}
    U_0 = \{( \mathbf{s}^{(l)}, \mathbf{0})_{l=1}^{L+1} : \mathbf{s}^{(l)} \in \mathbb{R}^{d_l}, \quad l=1,\dots, L+1    \}.
\end{align}

Since the network parameters $\mathbf{\theta}$ are fixed, there exists a constant $C(\theta)$  such that
\begin{align}
    \|\mathbf{V}^{(l)}\| \leq \frac{\|\mathbf{W}^{(l)}\|}{\|\mathbf{s}^{(l)}\|} \leq \frac{C(\theta)}{\|\mathbf{s}^{(l)}\|},  \quad l=1, \dots, L+1.  
\end{align}
For any weight factorization $(\mathbf{s}^{(l)}, \mathbf{V}^{(l)})_{l=1}^{L+1}$, we can take $(\mathbf{s}^{(l)}, 0)_{l=1}^{L+1} \in U_*$. By the definition of distance between sets, we obtain
\begin{align}
     \text{dist}(U_{\mathbf{\theta}}, U_0) = \min_{\mathbf{x} \in U_{\mathbf{\theta}}, \mathbf{y} \in  U_*} \|\mathbf{x} - \mathbf{y}\|
     \leq \sqrt{\sum_{l=1}^{L+1} \|\mathbf{V}^{(l)} \|^2 }  \leq C(\theta) \sqrt{\sum_{l=1}^{L+1} \frac{1}{\|\mathbf{s}^{(l)}\|^2}  }.
\end{align}
Therefore, for any network parameters $\theta, \theta'$, taking $C=\max\{C(\theta), C(\theta')\}$ yields
\begin{align}
      \text{dist}(U_{\mathbf{\theta}}, U_{\mathbf{\theta}'}) \leq \text{dist}(U_{\mathbf{\theta}}, U_0) + \text{dist}(U_0, U_{\mathbf{\theta}'}) \leq 2C\sqrt{\sum_{l=1}^{L+1} \frac{1}{\|\mathbf{s}^{(l)}\|^2}  }.
\end{align}

For $l=1, \dots, L+1$, letting $\mathbf{s}^{(l)}$ go to infinity, we have
\begin{align}
     \text{dist}(U_{\mathbf{\theta}}, U_{\mathbf{\theta}'})  = 0
\end{align}

As a corollary, let $\theta$ denote a network initialization and $\theta_*$ be a proper local minimum, then there exists a weight factorization with large enough scale factors $\mathbf{s}$,
such that the distance between $\theta$ and $\theta_*$ can be arbitrarily small in the factorized parameter space. 
\end{proof}

A different way to examine the effect of the proposed weight factorization is by studying its associated gradient updates. Recall that a standard gradient descent update with a learning rate $\eta$ takes the form 
\begin{align}  
    \label{eq: gradient_update}
    \mathbf{w}^{(k,l)}_{n+1} &= \mathbf{w}^{(k,l)}_{n} - \eta \frac{\partial \mathcal{L}}{\partial \mathbf{w}^{(k,l)}_{n}}.
\end{align}
The following theorem derives the corresponding gradient descent update expressed in the original parameter space for models using the proposed weight factorization.

\begin{theorem} Under the weight factorization of \eqref{eq: neuron_fact}, the gradient descent update is given by
\begin{align}
    \label{eq: fact_gradient_update}
    \mathbf{w}^{(k, l)}_{n+1} = \mathbf{w}^{(k, l)}_{n} - \eta \left([s^{(k,l)}_n]^2 + \|  \mathbf{v}^{(k,l)}_n\|^2_2 \right)  \frac{\partial \mathcal{L}}{\partial \mathbf{w}^{(k,l)}_{n}}  + \mathcal{O}(\eta^2),
\end{align}
for $l=1, 2,\dots, L+1$ and $k=1, 2, \dots, d_l$.
\end{theorem}

By comparing \eqref{eq: gradient_update} and  \eqref{eq: fact_gradient_update}, we observe that the weight factorization $\mathbf{w} = s \cdot \mathbf{v}  $ re-scales the learning rate of $\mathbf{w}$ by a factor of $(s^2 + \|\mathbf{v}\|_2^2)$. Since $\mathbf{s}, \mathbf{v}$ are trainable parameters, this analysis suggests that this weight factorization effectively assigns a self-adaptive learning rate to each neuron in the network.

\begin{proof}
Suppose that  $f^{(k, l)}$ denotes $k$-th component of $\mathbf{f}^{(l)} \in \R^{d_l}$. Under the proposed weight factorization in \eqref{eq: neuron_fact}, differentiating the loss function $\mathcal{L}$ with respect to $\mathbf{w}^{k,l}$ and $s^{(k, l)}$, respectively, yields 
\begin{align}
     s^{(k,l)}_{n+1} &=  s^{(k,l)}_{n} - \eta \frac{\partial \mathcal{L}}{\partial   s^{(k,l)}_{n}} =  s^{(k,l)}_{n} - \eta \frac{\partial \mathcal{L}}{\partial f^{(k, l)}} \cdot 
     \mathbf{v}^{(k,l)}_n \cdot \mathbf{g}^{(l-1)}, \\
      \mathbf{v}^{(k,l)}_{n+1} &=   \mathbf{v}^{(k,l)}_{n} - \eta \frac{\partial \mathcal{L}}{\partial   \mathbf{v}^{(k,l)}_{n}} = \mathbf{v}^{(k,l)}_{n} - \eta s^{(k,l)}_n \frac{\partial \mathcal{L}}{\partial f^{(k, l)}} \cdot \mathbf{g}^{(l-1)}.
\end{align}
Note that
\begin{align}
    \frac{\partial \mathcal{L}}{\partial \mathbf{w}^{(k,l)}_{n}} = \frac{\partial \mathcal{L}}{\partial f^{(k, l)}} \cdot \mathbf{g}^{(l-1)},
\end{align}
and the update rule of $\mathbf{v}^{(k,l)}$ and $ s^{(k,l)}$ can be re-written as
\begin{align}
      s^{(k,l)}_{n+1} &=  s^{(k,l)}_{n} - \eta \mathbf{v}^{(k,l)}_n \cdot \frac{\partial \mathcal{L}}{\partial \mathbf{w}^{(k,l)}_{n}}, \\
       \mathbf{v}^{(k,l)}_{n+1} &=  \mathbf{v}^{(k,l)}_{n} - \eta s^{(k,l)}_n   \frac{\partial \mathcal{L}}{\partial \mathbf{w}^{(k,l)}_{n}}.
\end{align}
Since $\mathbf{w}^{(k, l)} = s^{(k, l)} \cdot \mathbf{v}^{(k, l)} $, the update rule of $\mathbf{w}^{(k, l)}$ is given by
\begin{align}
    \mathbf{w}^{(k, l)}_{n+1} = \mathbf{w}^{(k, l)}_{n} - \eta \left([s^{(k,l)}_n]^2 + \| \mathbf{v}^{(k,l)}_n\|^2_2  \right)  \frac{\partial \mathcal{L}}{\partial \mathbf{w}^{(k,l)}_{n}}  + \mathcal{O}(\eta^2).
\end{align}
\end{proof}

\section{PINNs can violate causality}
\label{appendix: causality}

To illustrate that PINNs can violate causality, we closely examine the minimization of the PDE residual loss $\mathcal{L}_r$ (see equation \ref{eq: loss_r}). Before doing so, let us introduce some notation for convenience. Suppose that  $0 = t_1 < t_2  < \cdots < t_{N_t} = T$ discretizes the temporal domain, and $\{\mathbf{x}_j\}_{j=1}^{N_x}$ discretizes the spatial domain $\Omega$.  Now, for a given collection of spatial locations  $\{\mathbf{x}_j\}_{j=1}^{N_x}$, we can define the temporal residual loss as 
\begin{align}
    \label{eq: temporal_residual}
    \mathcal{L}_r(t, \mathbf{\theta}) = \frac{1}{N_x} \sum_{j=1}^{N_x} | \frac{\partial \mathbf{u}_{\mathbf{\theta}}}{\partial t}(t, \mathbf{x}_j) + \mathcal{N}[\mathbf{u}_{\mathbf{\theta}}](t, \mathbf{x}_j) |^2,
\end{align}

For a specified set of parameters $\mathbf{\theta}$ and a short time interval $[t^*, t^* +\Delta t]$, the PDE residual loss $L_r(t^*, \theta)$ essentially measures the deviation from the solution of the corresponding PDE
\begin{align}
    \mathcal{N}[\mathbf{u}](t, x) &= 0, \quad (t, x) \in [t^*, t^* + \Delta t] \times \Omega \\
    u(t^*, x) &= u_\theta(t^*, x), \quad x \in \Omega
\end{align}
Here, $u_\theta(t^*, x)$ represents the network's prediction using the given fixed parameters $\theta$ evaluated at $t^*$. As a result, even if $\mathcal{L}_r(t^*, \theta)=0$, the accuracy of the predicted solution in $[t^*, t^* + \Delta t]$ is determined by the deviation of $u_\theta(t^*, x)$ from the ground truth and thus the error will propagate alone the time. Hence, we argue that  the temporal residual loss $L_r(t, \theta)$ should be based on the current predicted solution at time $t$ and the optimization process is meaningful only if the predicted solution is reasonable for previous times.

Note that the residual loss \ref{eq: loss_r} can be rewritten as 
\begin{align}
    \mathcal{L}_r(\mathbf{\theta}) &=   \frac{1}{N_t} \sum_{i=1}^{N_t} \mathcal{L}_r(t_i, \mathbf{\theta})  \\
    &=  \frac{1}{N_t N_x} \sum_{i=1}^{N_t} \sum_{j=1}^{N_x} | \frac{\partial \mathbf{u}_{\mathbf{\theta}}}{\partial t}(t_i, \mathbf{x}_j) + \mathcal{N}[\mathbf{u}_{\mathbf{\theta}}](t_i, \mathbf{x}_j) |^2. 
\end{align}

Next, we discretize $\frac{\partial \mathbf{u}_{\mathbf{\theta}}}{\partial t}$ using the forward Euler scheme \cite{iserles2009first}. For any $1 \leq i \leq N_t -1$, $\mathcal{L}(t_i, \mathbf{\theta}) $  can be approximated by
\begin{align}
    \mathcal{L}_r(t_i, \mathbf{\theta})  &\approx  \frac{1}{N_x} \sum_{j=1}^{N_x} \left|  \frac{\mathbf{u}_{\mathbf{\theta}}(t_{i}, \mathbf{x}_j) - \mathbf{u}_{\mathbf{\theta}}(t_{i-1}, \mathbf{x}_j)}{\Delta t}  + \mathcal{N}[\mathbf{u}_{\mathbf{\theta}}](t_{i}, \mathbf{x}_j) \right
    |^2 \nonumber \\
    &\approx \frac{|\Omega|}{\Delta t^2} \int_\Omega |\mathbf{u}_{\mathbf{\theta}}(t_{i}, \mathbf{x}) - \mathbf{u}_{\mathbf{\theta}}(t_{i-1}, \mathbf{x}) + \Delta t \mathcal{N}[\mathbf{u}_{\mathbf{\theta}}](t_{i}, \mathbf{x})   |^2 d\mathbf{x}.
\end{align}
From the above expression, we immediately obtain that the proper minimization of $\mathcal{L}_r(t_i, \mathbf{\theta})$ should be based on the correct prediction of both  $\mathbf{u}_{\mathbf{\theta}}(t_{i}, \mathbf{x})$ and $\mathbf{u}_{\mathbf{\theta}}(t_{i-1}, \mathbf{x})$,  while the original formulation
tends to minimize all $\mathcal{L}_r(t_i, \mathbf{\theta})$ simultaneously. As a result,  the residual loss $\mathcal{L}_r(t_i, \mathbf{\theta})$ will be minimized even if the predictions at $t_i$ and previous times are inaccurate. 
This behavior inevitably violates temporal causality, making the PINN model susceptible to learn erroneous solutions.

\begin{table}[h]
\renewcommand{\arraystretch}{1.2}
\centering
\caption{{\em Allen-Cahn equation:} Hyper-parameter configuration.}
\label{tab: ac_config}
\begin{tabular}{ll}
\toprule
\textbf{Parameter} & \textbf{Value} \\
\midrule
\textbf{Architecture Parameters} & \\
Architecture & Modified MLP \\
Number of layers & 4 \\
Layer size & 256 \\
Activation & Tanh \\
Fourier feature scale & 2.0 \\
RWF & $\mu =0.5, \sigma=0.1$ \\
\addlinespace  
\textbf{Training Parameters} & \\
Learning rate & $0.001$ \\
Decay steps & 5,000 \\
Training steps & 300,000 \\
Batch size & 8,192 \\
\addlinespace  
\textbf{Weighting Parameters} & \\
Weighting scheme & NTK \\
Causal tolerance & 1.0 \\
Number of chunks & 32 \\
\bottomrule
\end{tabular}
\end{table}

\begin{figure}[h]
    \centering
    \includegraphics[width=0.8\textwidth]{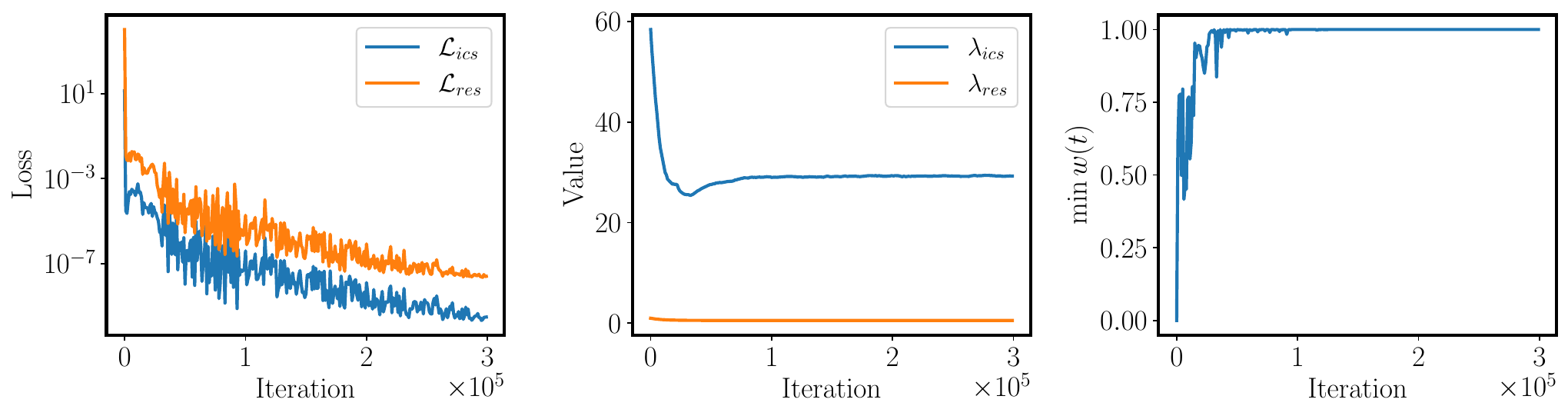}
        \caption{{\em Allen-Cahn equation:} {\em left:} Loss convergence of the initial condition loss $\mathcal{L}_{ics}$ and the PDE residual loss $\mathcal{L}_{res}$ during training. {\em Mid:} Changes of loss weights during training. {\em Right:} Minimum value of the temporal residual $\min_t w(t) during training$. All temporal PDE residuals are properly minimized if  $\min_t w(t) $ converge to 1.}
    \label{fig: ac_loss_weights}
\end{figure}

\begin{table}[h]
\renewcommand{\arraystretch}{1.2}
\centering
\caption{{\em Advection equation:} Hyper-parameter configuration.}
\label{tab: adv_config}
\begin{tabular}{ll}
\toprule
\textbf{Parameter} & \textbf{Value} \\
\midrule
\textbf{Architecture Parameters} & \\
Architecture & Modified MLP \\
Number of layers & 4 \\
Layer size & 256 \\
Activation & Tanh \\
Fourier feature scale & 1.0 \\
RWF & $\mu =1.0, \sigma=0.1$ \\
\addlinespace  
\textbf{Training Parameters} & \\
Learning rate & $0.001$ \\
Decay steps & 2,000 \\
Training steps & 200,000 \\
Batch size & 8,192 \\
\addlinespace  
\textbf{Weighting Parameters} & \\
Weighting scheme & Grad Norm \\
Causal tolerance & 1.0 \\
Number of chunks & 32 \\
\bottomrule
\end{tabular}
\end{table}

\begin{figure}[h]
    \centering
    \includegraphics[width=0.8\textwidth]{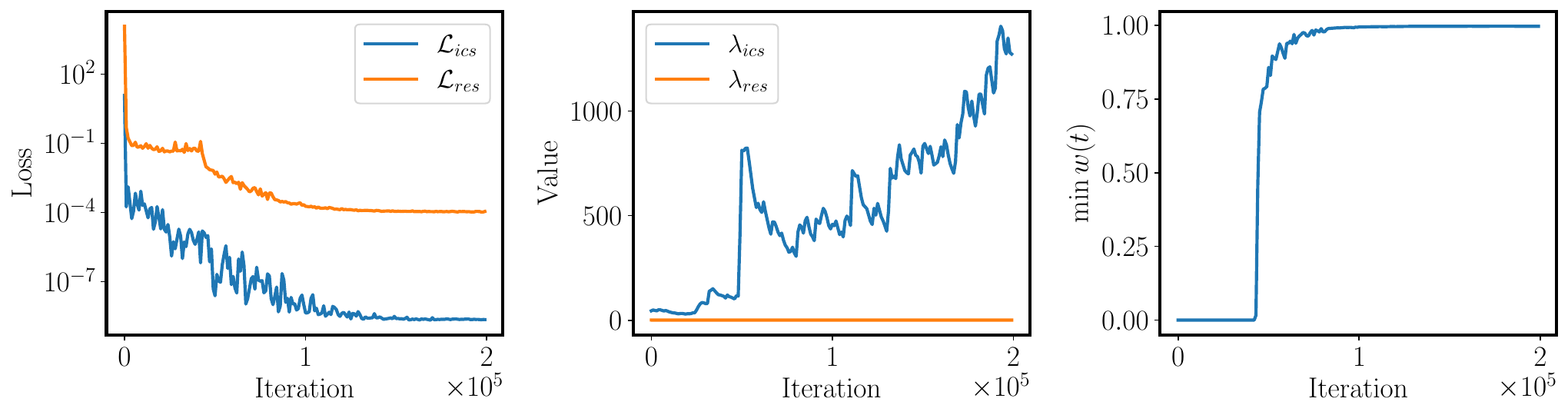}
    \caption{{\em Advection equation:} {\em left:} Loss convergence of the initial condition loss $\mathcal{L}_{ics}$ and the PDE residual loss $\mathcal{L}_{res}$ during training. {\em Mid:} Changes of loss weights during training. {\em Right:} Minimum value of the temporal residual $\min_t w(t) during training$. All temporal PDE residuals are properly minimized if  $\min_t w(t) $ converge to 1.}
    \label{fig: adv_loss_weights}
\end{figure}

\begin{table}[h]
\renewcommand{\arraystretch}{1.2}
\centering
\caption{{\em  Stokes equation:} Hyper-parameter configuration.}
\label{tab: stokes_config}
\begin{tabular}{ll}
\toprule
\textbf{Parameter} & \textbf{Value} \\
\midrule
\textbf{Architecture Parameters} & \\
Architecture & Modified MLP \\
Number of layers & 4 \\
Layer size & 256 \\
Activation & GeLU \\
Fourier feature scale & 10.0 \\
RWF & $\mu =0.5, \sigma=0.1$ \\
\addlinespace  
\textbf{Training Parameters} & \\
Learning rate & $0.001$ \\
Decay steps & 2,000 \\
Training steps & 100,000 \\
Batch size & 8,192 \\
\addlinespace  
\textbf{Weighting Parameters} & \\
Weighting scheme & Grad Norm \\
\bottomrule
\end{tabular}
\end{table}

\begin{figure}[h]
    \centering
    \includegraphics[width=0.8\textwidth]{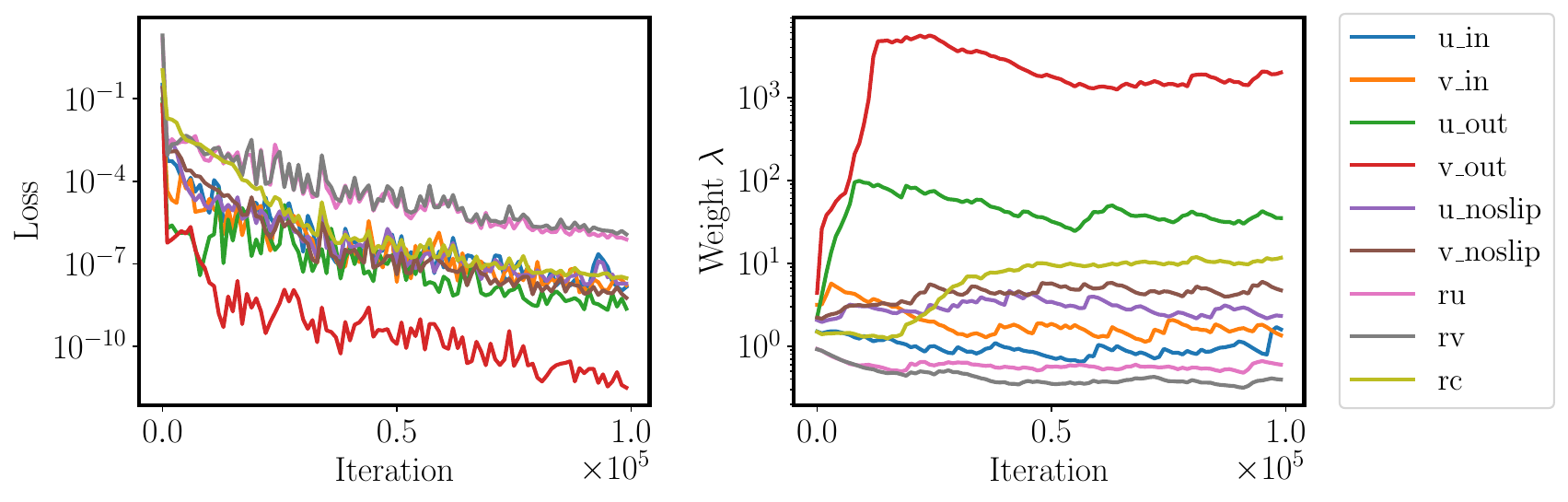}
    \caption{{\em Stokes equation:} {\em left:} Loss convergence of the initial condition loss $\mathcal{L}_{ics}$ and the PDE residual loss $\mathcal{L}_{res}$ during training. {\em Right:} Changes of loss weights during training.}
    \label{fig: stokes_loss_weight}
\end{figure}

\begin{table}[h]
\renewcommand{\arraystretch}{1.2}
\centering
\caption{{\em  Kuramoto–Sivashinsky equation:} Hyper-parameter configuration.}
\label{tab: ks_config}
\begin{tabular}{ll}
\toprule
\textbf{Parameter} & \textbf{Value} \\
\midrule
\textbf{Architecture Parameters} & \\
Architecture & Modified MLP \\
Number of layers & 5 \\
Layer size & 256 \\
Activation & Tanh \\
Fourier feature scale & 1.0 \\
RWF & $\mu =0.5, \sigma=0.1$ \\
\addlinespace  
\textbf{Training Parameters} & \\
Learning rate & $0.001$ \\
Decay steps & 2,000 \\
Number of time windows & 10 \\
Training steps per window & 200,000 \\
Batch size & 4,096 \\
\addlinespace  
\textbf{Weighting Parameters} & \\
Weighting scheme & Grad Norm \\
Causal tolerance & 1.0 \\
Number of chunks & 16 \\
\bottomrule
\end{tabular}
\end{table}

\begin{figure}[h]
    \centering
    \includegraphics[width=0.8\textwidth]{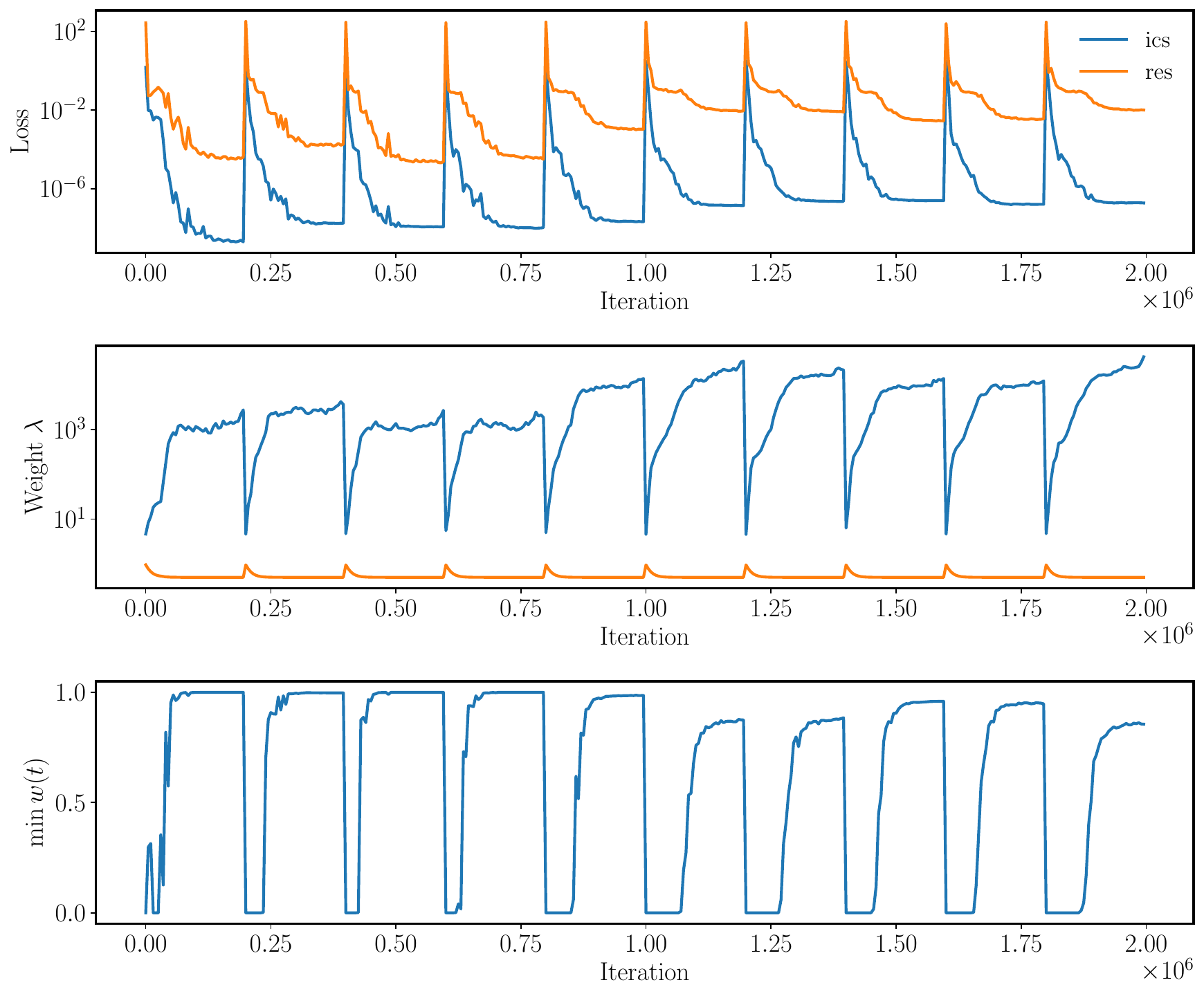}
    \caption{{\em Kuramoto–Sivashinsk equation:} {\em Top:} Loss convergence of the initial condition loss $\mathcal{L}_{ics}$ and the PDE residual loss $\mathcal{L}_{res}$ during training. {\em Mid:} Changes of loss weights during training. {\em Bottom:} Changes of causal weights during training.} 
    \label{fig: ks_loss_weight}
\end{figure}

\begin{table}[h]
\renewcommand{\arraystretch}{1.2}

\centering
\caption{{\em Lid-driven cavity flow:} Hyper-parameter configuration.}
\label{tab: ldc_config}
\begin{tabular}{ll}
\toprule
\textbf{Parameter} & \textbf{Value} \\
\midrule
\textbf{Architecture Parameters} & \\
Architecture & Modified MLP \\
Number of layers & 5 \\
Layer size & 256 \\
Activation & Tanh \\
Fourier feature scale & 10.0 \\
RWF & $\mu =1.0, \sigma=0.1$ \\
\addlinespace  
\textbf{Training Parameters} & \\
Learning rate & $0.001$ \\
Decay steps & 10,000 \\
Batch size & 8,192 \\
\addlinespace  
\textbf{Curriculum Training} & \\
$Re$ & [100, 400, 1,000, 3,200] \\
Training steps & [50,000, 50,000, 10,0000, 50,0000] \\
\addlinespace  
\textbf{Weighting Parameters} & \\
Weighting scheme & Grad Norm \\
\bottomrule
\end{tabular}
\end{table}

\begin{table}[]
    \centering
    \renewcommand{\arraystretch}{1.2}
    \begin{tabular}{ccccccc}
    \toprule
    \multicolumn{4}{c}{\textbf{Ablation Settings}} & \multicolumn{2}{c}{\textbf{Performance}} \\ 
    \cmidrule(lr){1-4} \cmidrule(lr){5-6}
    \textbf{Fourier Feature} & \textbf{RWF} & \textbf{Grad Norm} & \textbf{Modified MLP} & \textbf{Rel. $L^2$ error} & \textbf{Run time (min) } \\ 
    \midrule
    $\cmark$ & $\cmark$ & $\cmark$ & $\cmark$ & $\mathbf{1.34 \times 10^{-1}}$ & 58.86  \\ 
    $\xmark$ & $\cmark$ & $\cmark$ & $\cmark$ &  $7.32 \times 10^{-1}$ &  51.28 \\ 
    $\cmark$ & $\xmark$ & $\cmark$ & $\cmark$ &  $1.59 \times 10^{-1}$ &  62.01 \\ 
    $\cmark$ & $\cmark$ & $\xmark$ & $\cmark$ &  $3.38 \times 10^{-1}$ &  57.16 \\ 
    $\cmark$ & $\cmark$ & $\cmark$ & $\xmark$ &  $5.48 \times 10^{-1}$ &  23.40 \\ 
    $\xmark$ & $\xmark$ & $\xmark$ & $\xmark$ &  $7.94 \times 10^{-1}$ &  17.96 \\ 
    \bottomrule
    \end{tabular}
    \caption{{\em Lid-driven cavity flow:}   Relative $L^2$ error and run time for an ablation study illustrating the impact of disabling nondimensionalization and individual components of the proposed training pipeline. The error is measured against  the norm of flow velocity $\|\mathbf{u}\|_2 = \sqrt{u^2 + v^2}$  The first row corresponds to the fine-tuned model using Modified MLP whose hyper-parameter configuration can be found in Table \ref{tab: ldc_config}. }
    \label{tab: ldc}
\end{table}

\begin{figure}
    \centering
    \includegraphics[width=0.5\textwidth]{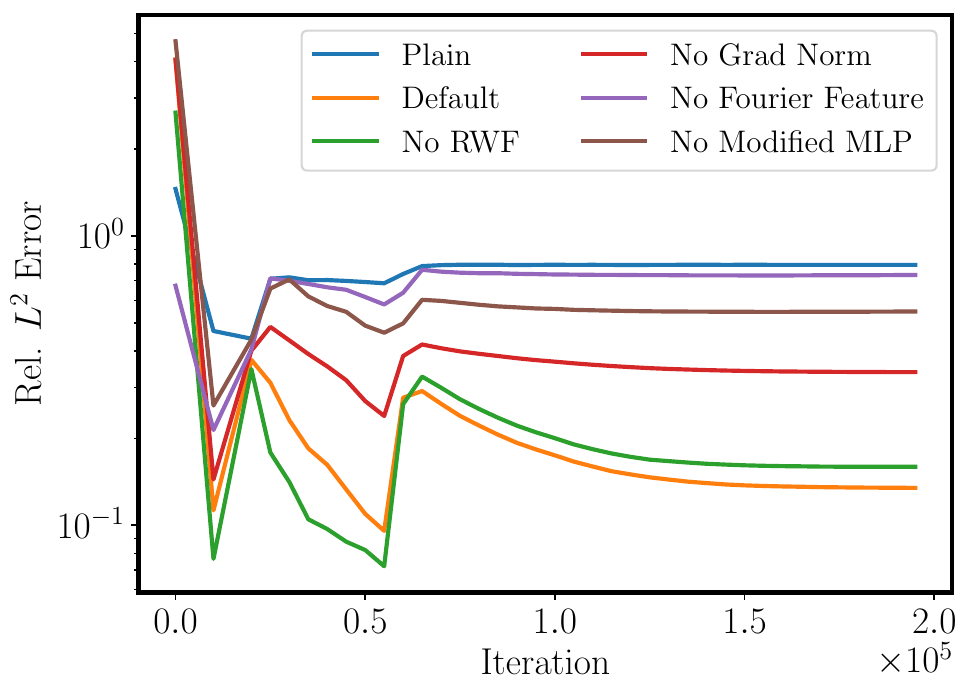}
    \caption{{\em Lid-driven cavity:} Convergence of relative $L^2$ error for the ablation study with different components disabled. }
    \label{fig: ldc_ablation}
\end{figure}

\begin{table}[h]
\renewcommand{\arraystretch}{1.2}
\centering
\caption{{\em  Navier-Stokes flow in torus:} Hyper-parameter configuration.}
\label{tab: ns_tori_config}
\begin{tabular}{ll}
\toprule
\textbf{Parameter} & \textbf{Value} \\
\midrule
\textbf{Architecture Parameters} & \\
Architecture & Modified MLP \\
Number of layers & 4 \\
Layer size & 256 \\
Activation & Tanh \\
Fourier feature scale & 1.0 \\
RWF & $\mu =0.5, \sigma=0.1$ \\
\addlinespace  
\textbf{Training Parameters} & \\
Learning rate & $0.001$ \\
Decay steps & 2,000 \\
Number of time windows & 5 \\
Training steps per window & 150,000 \\
Batch size & 8,192 \\
\addlinespace  
\textbf{Weighting Parameters} & \\
Weighting scheme & Grad Norm \\
Causal tolerance & 1.0 \\
Number of chunks & 16 \\
\bottomrule
\end{tabular}
\end{table}

\begin{figure}[h]
    \centering
    \includegraphics[width=0.8\textwidth]{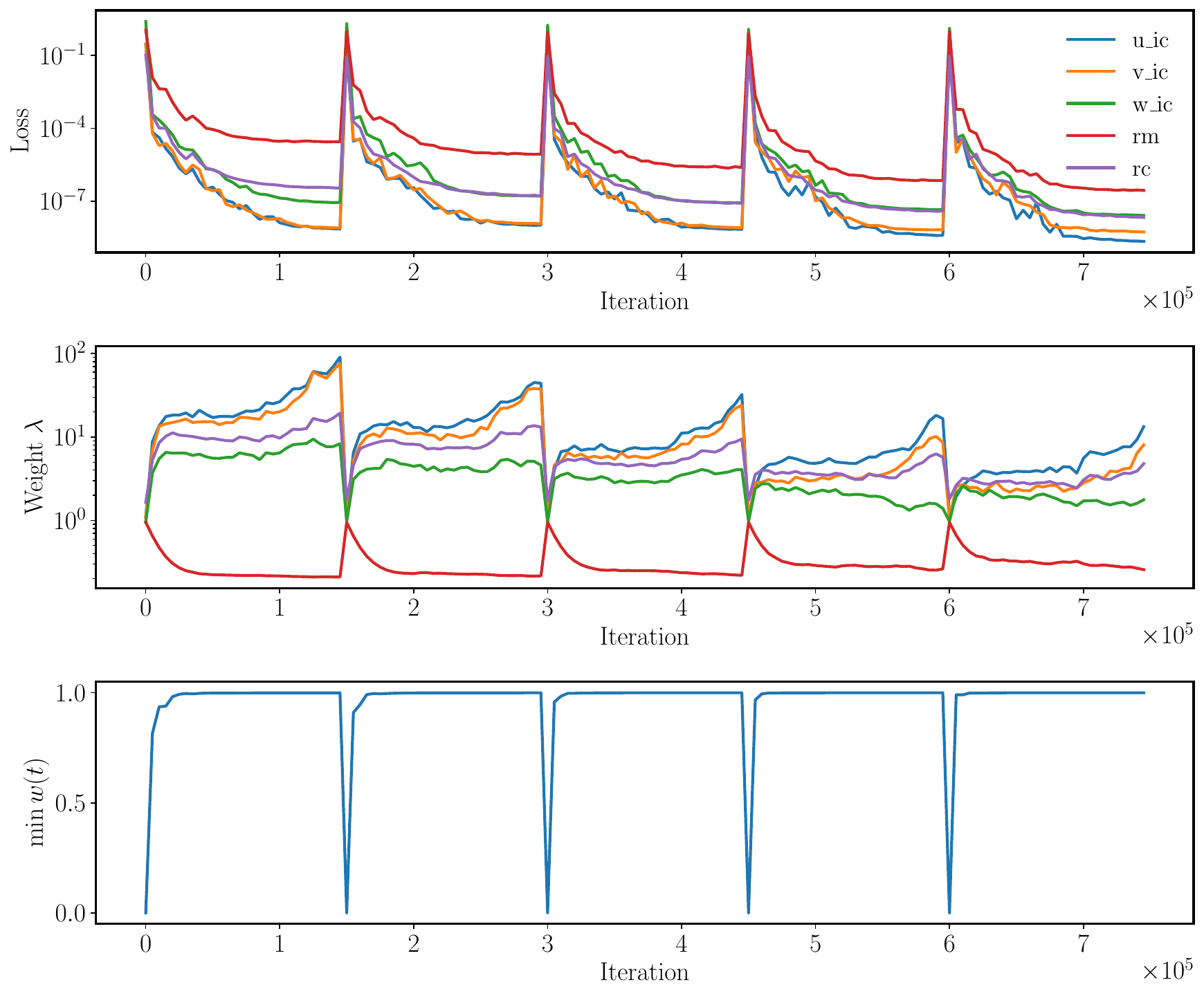}
    \caption{{\em Navier-Stokes flow in torus:} {\em Top:} Loss convergence of the initial condition loss $\mathcal{L}_{ics}$ and the PDE residual loss $\mathcal{L}_{res}$ during training. {\em Mid:} Changes of loss weights during training. {\em Bottom:} Changes of causal weights during training.} 
    \label{fig: ns_tori_loss_weight}
\end{figure}

\begin{table}[h]
\renewcommand{\arraystretch}{1.2}
\centering
\caption{{\em  Navier-Stokes flow around a cylinder:} Hyper-parameter configuration.}
\label{tab: ns_cylinder_config}
\begin{tabular}{ll}
\toprule
\textbf{Parameter} & \textbf{Value} \\
\midrule
\textbf{Architecture Parameters} & \\
Architecture & Modified MLP \\
Number of layers & 5 \\
Layer size & 256 \\
Activation & Tanh \\
Fourier feature scale & 1.0 \\
RWF & $\mu =1.0, \sigma=0.1$ \\
\addlinespace  
\textbf{Training Parameters} & \\
Learning rate & $0.001$ \\
Decay steps & 2,000 \\
Number of time windows & 10 \\
Training steps per window & 200,000 \\
Batch size & 8,192 \\
\addlinespace  
\textbf{Weighting Parameters} & \\
Weighting scheme & Grad Norm \\
Causal tolerance & 1.0 \\
Number of chunks & 16 \\
\bottomrule
\end{tabular}
\end{table}

\begin{figure}[h]
    \centering
    \includegraphics[width=0.8\textwidth]{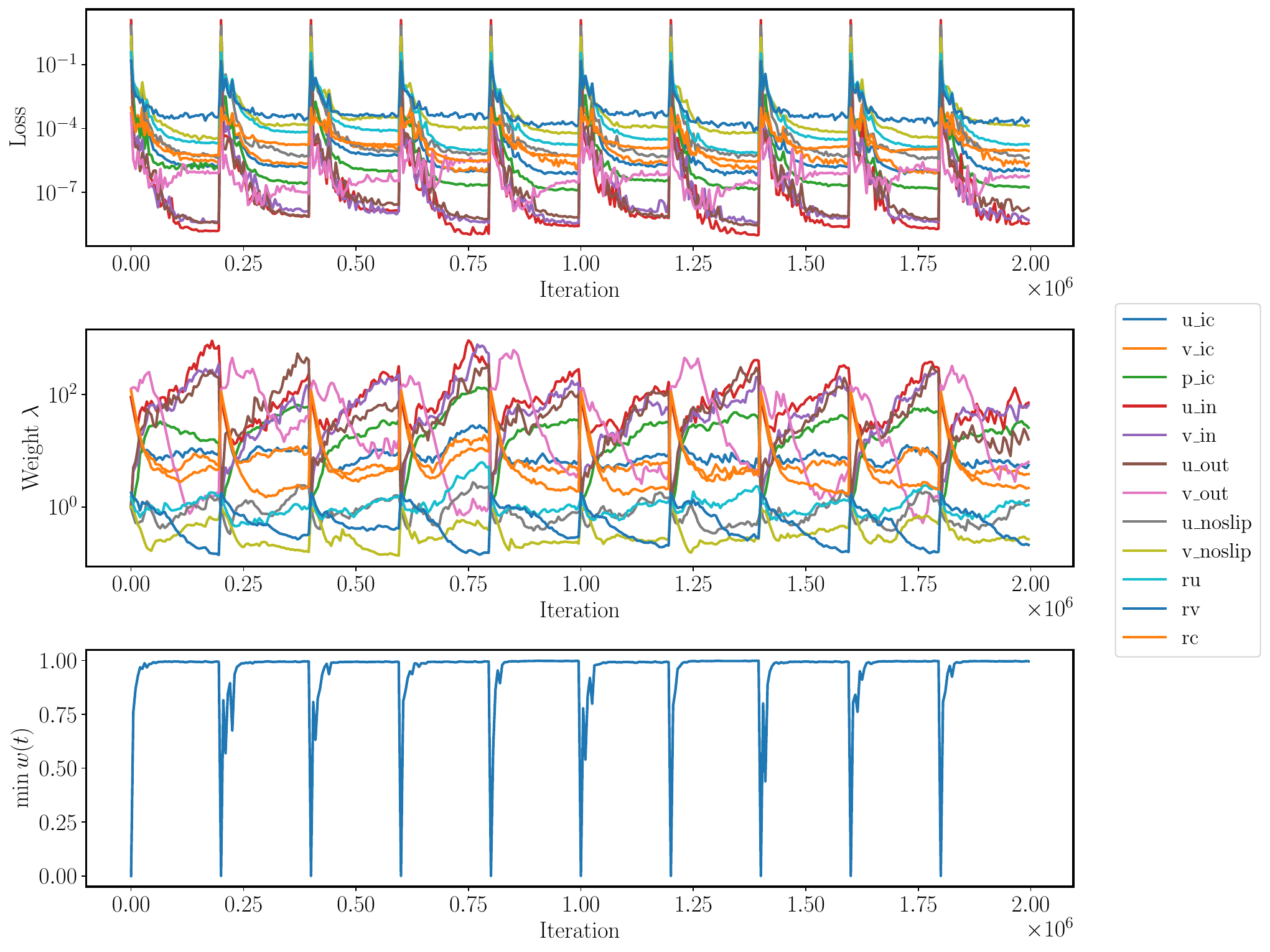}
    \caption{{\em Navier-Stokes flow around cylinder:} {\em Top:} Loss convergence of the initial condition loss $\mathcal{L}_{ics}$ and the PDE residual loss $\mathcal{L}_{res}$ during training. {\em Mid:} Changes of loss weights during training. {\em Bottom:} Changes of causal weights during training.} 
    \label{fig: cylinder_loss_weight}
\end{figure}

\end{document}